\documentclass[12pt, a4paper]{article}

\usepackage[a4paper, top=2.5cm, bottom=2.5cm, left=2.5cm, right=2.5cm]{geometry}
\usepackage{amsmath}
\usepackage{amssymb}
\usepackage{hyperref} % Keep this last for hyperlinking in the PDF
\usepackage{booktabs}
\usepackage{graphicx} % To use the \centering command
\usepackage{color}
\usepackage{authblk}
\usepackage[numbers,sort&compress]{natbib}
\newcommand{\DNN}{Deep Neural Network}
\newcommand{\LIF}{Leaky Integrate-and-Fire}

\newcommand{\rem}[1]{}

\RequirePackage[normalem]{ulem} %DIF PREAMBLE
\RequirePackage{color}\definecolor{RED}{rgb}{1,0,0}\definecolor{BLUE}{rgb}{0,0,1} %DIF PREAMBLE
\title{The Neural Division of Labor: Biologically-Inspired Modular Architectures for Robust Neuromorphic Computing}
\author[1]{Maxim Bazhenov}
\author[2]{Serafim Grubas}
\author[3,4,$\dagger$]{Vakhtang Putkaradze\thanks{Professor Emeritus at the University of Alberta, where the initial part of this work was performed}  }

\affil[1]{Department of Medicine, University of California San Diego, La Jolla, CA 92093, USA}
\affil[2]{Department of Physics, University of Alberta, Edmonton, AB T6G 2G1, Canada}
\affil[3]{Department of Mathematical and Statistical Sciences, University of Alberta, Edmonton, AB T6G 2G1, Canada}
\affil[4]{Department of Mathematics, University of Alabama, Tuscaloosa, AL 35487, USA}
\affil[$\dagger$]{Corresponding author: vputkaradze@ua.edu}

\begin{document}

\maketitle

% Footnote for Emeritus status
\renewcommand{\thefootnote}{\fnsymbol{footnote}}
\renewcommand{\thefootnote}{\arabic{footnote}}

\maketitle
\thispagestyle{empty} % Suppress page number on the title page (common practice)

\begin{abstract}
Biological neural systems achieve high efficiency and robustness through compartmentalized architectures. In contrast, modern artificial neural networks rely on globally entangled structures, which obscure decision logic and suffer from catastrophic forgetting. Here, we report a Decomposable Spiking Neural Network (D-SNN) that eliminates global synaptic entanglement by structurally isolating classification pathways into independent experts. Optimized via a bio-inspired push-pull loss function, the D-SNN achieves competitive accuracies on MNIST, Fashion-MNIST, and CIFAR-10/100 benchmarks. This modular approach matches the performance of fully dense networks while utilizing an order of magnitude fewer parameters. In addition, our networks operate with up to several orders of magnitude lower firing rates and fewer synaptic operations. Furthermore, physically severing connections between experts provides inherent protection against catastrophic forgetting during sequential learning. Crucially, these isolated pathways generate auditable neural signals, increasing decision transparency. This biomimetic, verifiable architecture establishes an efficient foundation for deploying deterministic neuromorphic intelligence in resource-constrained edge environments.
\end{abstract}

\section{Introduction} \label{sec:intro}

Artificial Intelligence (AI), driven primarily by Artificial Neural Networks (ANNs), has revolutionized domains ranging from image classification \cite{krizhevsky2012imagenet, lecun2015deep} to healthcare diagnostics \cite{topol2019high}, security systems \cite{morgan2020military}, and wireless communications \cite{oshea2017introduction}. However, as AI scales to massive architectures like Large Language Models \cite{brown2020language, achiam2023gpt, kaplan2020scaling}, its energy trajectory has become unsustainable. The computational power required for state-of-the-art AI doubles approximately every 3.4 months, vastly outpacing hardware efficiency gains \cite{OpenAI2018, Dhar2020, Desislavov2023}. Consequently, AI data centers already consume equivalent to several percent of global energy \cite{IEA2023}, a figure projected to reach up to 12\% of US electricity consumption by 2028 \cite{Shehabi2024}. Concurrently, massive parameter counts create ``black box'' models, obscuring decision logic and making it impossible to guarantee safety in critical failure modes \cite{lecun2015deep, rudin2019stop}. While post-hoc explainability techniques (e.g. concept bottlenecks \cite{koh2020concept, chauhan2023interactive}, feature learning \cite{bricken2023monosemanticity}, and gradient attribution \cite{sundararajan2017axiomatic}) offer insights, they remain difficult to interpret directly. On the contrary, developing models that are interpretable by design can offer competitive accuracy and with greater transparency for audit in high-stakes decisions, but they often require more effort to develop \cite{rudin2019stop}.

Neuromorphic computing offers a promising path to energy efficiency  by mimicking biological, event-driven processing \cite{pei2019towards, debole2019truenorth, davies2018loihi, fang2021incorporating, zenke2021brain, yao2024spike}. Spiking Neural Networks (SNNs) can theoretically achieve energy efficiency orders of magnitude greater than traditional von Neumann architectures \cite{cao2015spiking, roy2019towards, kosters2023benchmarking, abreu2025neuromorphic, imanov2026neuedge}, provided their activations remain highly sparse. However, natively training SNNs is notoriously resource-intensive due to the non-differentiable, time-dependent nature of spiking dynamics. Methods like Backpropagation Through Time (BPTT) \cite{werbos1990backpropagation, guo2023efficient}, surrogate gradients \cite{neftci2019surrogate}, and eligibility propagation \cite{zenke2021remarkable, eshraghian2023training} are computationally heavy. Conversely, ANN-to-SNN weight conversion often sacrifices stability and structural sparsity \cite{bu2023optimal, tavanaei2019deep, wu2018spatio}. Furthermore, the temporal dynamics of SNNs make interpretability exceedingly difficult, prompting ongoing research into feature tracking \cite{nguyen2023feature}, temporal gradient attribution \cite{bitar2023gradient}, causal models \cite{kar2025binary}, and deep learning dynamics \cite{kasabov2021deep}.

A way to the model interpretability is the One-vs-All (OvA) and One-vs-One (OvO) classification architectures \cite{rifkin2004defense, pawara2020one}, where decision-making is partitioned among independent experts. In such architectures, once a classification decision is made, its origin is unequivocally clear. Despite these benefits, generating and training OvA networks in the spiking domain has proven difficult. To overcome this, we draw inspiration from the decentralized and modular nature of biological nervous systems and propose a Decomposable Spiking Neural Network (D-SNN). This architecture explicitly prioritizes biological plausibility, functional sparsity, energy efficiency, and formal explainability. Central to our approach is a novel biomimetic ``Push-Pull'' loss function. Much like the Antennal Lobe (AL) inhibitory networks found in biological olfaction, our optimization aggressively suppresses shared representations (crosstalk) and enhances distinct, class-specific features. Only the target expert pathway receives the synaptic input required to overcome the threshold in the leaky integrate-and-fire (LIF) model of a spiking neuron, ensuring that non-target experts remain metabolically quiescent.

This mechanism creates $N$ disjoint SNN experts, each dedicated to a One-vs-All classification task \cite{rifkin2004defense}. When coupled with our biomimetic Push-Pull training procedure, enforcing the neural activity in correct experts, D-SNN is substantially more transparent for verification of results and identifying which combination of experts was responsible for the solution. 

Our benchmarks show that our D-SNN models achieve over 98\% accuracy on MNIST and 90\% on Fashion-MNIST, maintaining competitive performance on CIFAR-10 at $\sim66$\%, while using an order of magnitude fewer parameters than leading deep learning models. Moreover, the D-SNNs resist catastrophic forgetting and provide classification results characterized by activation sparsity and high level of metabolic efficiency.

%%%% END REM 

% While the extreme efficiency of insect sensory systems is often attributed to purely innate, hardwired architectures, recent biological evidence demonstrates that these networks utilize sophisticated, localized plasticity to dynamically develop highly structured, non-overlapping forward pathways. For instance, in the honeybee olfactory system, the early sensory relay known as the antennal lobe (AL) must learn to distinguish complex, overlapping odor mixtures to identify rewards \cite{smith2011distributed,lei2022novelty,joshi2025plasticity}. Recent in vivo and computational studies reveal that training induces inhibitory plasticity within the AL circuits; the network actively suppresses responses to shared chemical compounds while enhancing responses to distinct, reward-specific features. This active remodeling results in profound pattern separation and the emergence of a highly concise, separated neural code tailored to the environment. The Decomposable Spiking Neural Network (D-SNN) explicitly mirrors this biological strategy. Much like the AL inhibitory network, our biomimetic 'Push-Pull' optimization aggressively suppresses shared representations (crosstalk) and enhances distinct class features. This process actively carves out the non-overlapping, structurally isolated expert pathways that grant the D-SNN its remarkable metabolic efficiency and resistance to capacity collapse.

\section{Results}
%\subsection{Model architecture}

%We employ training of our network through a specially designed Artifical Neural Network (ANN) first. The weights of learned ANN with a specially designed loss function are then imported directly into a spiking network with exactly the same architecture; no learning of SNN is undertaken. This procedure allows to achieve a radical increase in efficiency and speed of learning procedure compared to learning a full-fledged SNN, with equivalent accuracy results. However, our SNN allows for radical improvement in explanability and traceabilty of results. 
\subsection{Model Architecture and Training Methodology}
\begin{figure}
    \centering
    \includegraphics[width=1\linewidth]{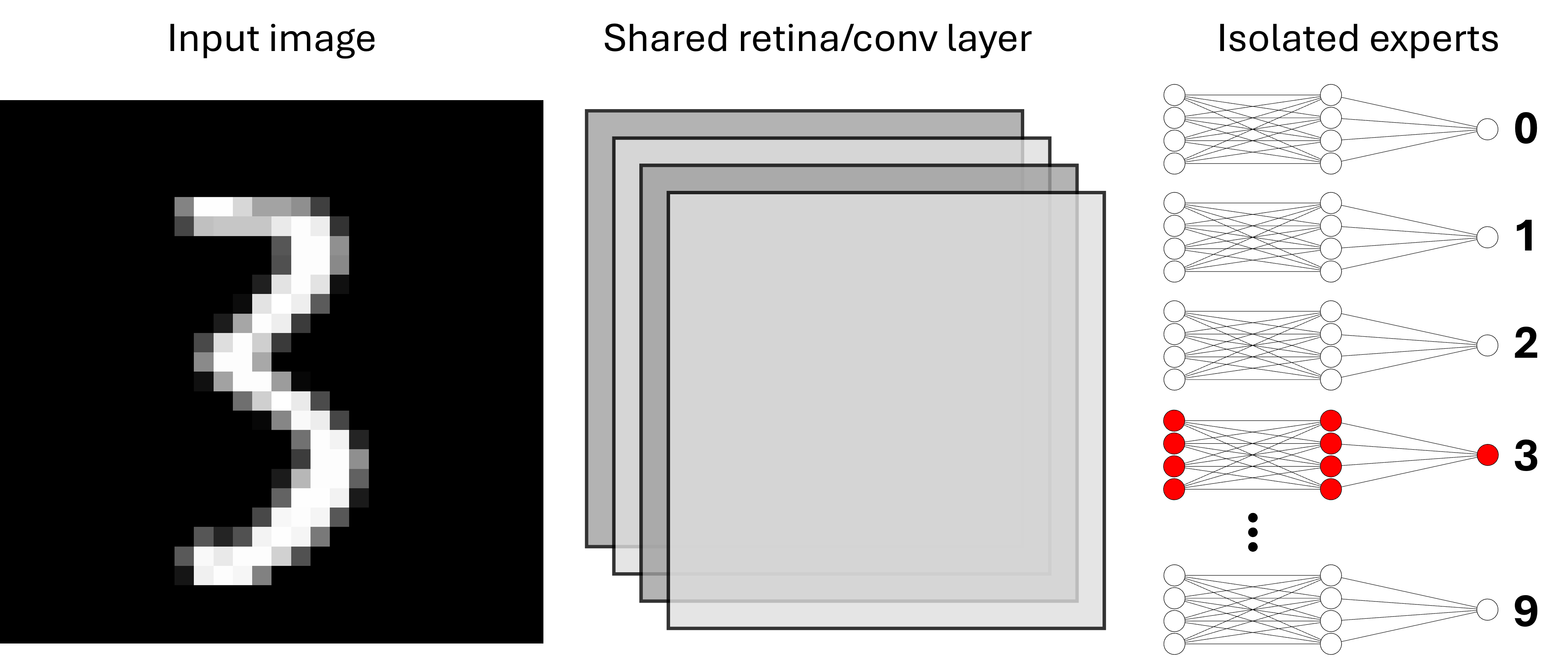}
    \caption{\textbf{Decomposable sensory architecture enabling parallel class verification in the D-SNN.} 
In analogy of decentralized and specialized nervous systems found in insects, our Decomposable Spiking Neural Network (D-SNN) is designed with explicit structural modularity. Left panel: A raw visual input signal, illustrated by an MNIST digit '3'. Center panel: The signal is first processed by a tiered layers of convolutional layers, to retinal and early cortical processing.
Right panel: The key innovation of this paper lies in the method of training of hidden processing layers, which are structurally partitioned into a series of discrete, isolated functional units ('experts'), representing specialist pathways, labeled 0--9 in this case. Each pathway is a compact, self-contained spiking circuit dedicated to verifying a single output class via a One-vs-All framework \cite{rifkin2004defense}. This compartmentalization ensures that visual decisions are inherently interpretable and formally verifiable, preventing signal interference across disparate tasks. When presented with an input from class '3', our biomimetic training objective forces the total neural activity to localize within the target specialist pathway, highlighted in red, while all other pathways remain effectively quiescent (white). 
%This approach provides a blueprint for creating functionally robust, verifiable robotic controllers with extreme energy sparsity.
    }
    \label{fig:experts}
\end{figure}

We structure the proposed network into two main components: a shared feature extraction layers and a set of either dense or partitioned hidden layers. The feature extraction front end consists of standard convolutional layers that extract continuous spatial features from the input data. Following this backbone, we consider three different cases. The first two cases represent the standard architecture of two dense layers, with the only difference being the way of training these hidden layers, as described below in Section~\ref{sec:biomiectic_loss}. Depending on the loss function used, we call these cases \emph{Dense Hybrid} when the new loss function \eqref{loss_def_0} is used, and \emph{Dense CCE} when the standard Categorical Cross-Entropy (CCE) is used for training. In the third case, we partition the hidden layers into $K$ discrete, non-overlapping functional units, or \textit{experts}, where $K$ corresponds to the number of output classes. We call this case \emph{Independent Experts}. In that case, each expert pathway is allocated a disjoint subset of hidden neurons connecting exclusively to a single output neuron representing its target class, as illustrated in Figure~\ref{fig:experts}; cross-talk between experts is explicitly prohibited by zeroing out the weight matrices beyond block-diagonal elements. This structural modularity mirrors the specialized, low-redundancy neural pathways found in biological sensory systems, such as the insect olfactory circuit \cite{joshi2025plasticity}, and ensures that each expert processes class-specific features independently.

We then train the network entirely within a continuous artificial neural network (ANN) framework using a specially designed biomimetic loss function \eqref{loss_def_0} in Section~\ref{sec:biomiectic_loss} for the independent experts and dense hybrid cases, and standard CCE loss function for the dense CCE case. Once the training converges, the learned weights are transferred directly into a spiking neural network (SNN) of identical architecture, with no further learning performed in the spiking domain. This two-stage approach, when the training occurs in ANN, and all verification is done in SNN,  achieves the same classification accuracy as direct SNN training while drastically reducing computational cost, since it avoids the complexity of optimizing discrete, non-differentiable spiking dynamics.

%The ANN, and the corresponding SNN, consists of standard one- or two-layer convolution layers, and a specially designed SNN constructed from  decomposable layers. To implement the One-vs-All expert architecture, we partition the hidden layers of the network into $K$ discrete functional units, or 'experts' each corresponding to a specific output class. This structural modularity mimics the specialized neural paths found in biological sensory systems. Thus, each path creates an expert in a particular class. 

%By measuring the cumulative neural output of these specialized clusters, we observed that the network autonomously learned to isolate class-specific features. 

The partitioning effectively decoupled the global dense network into a series of independent experts, providing the basis for parallel verification and the prevention of signal interference across disparate tasks. The resulting SNN, by virtue of its modular expert architecture, provides substantially improved interpretability and decision traceability compared to standard dense networks: because each class is processed by a physically isolated pathway, the origin of any classification decision can be directly and unambiguously identified from the network's activation pattern.

\subsection{Biomimetic loss function for competitive pathway reinforcement} 
\label{sec:biomiectic_loss}

%Once the architecture has been established, we perform the training. 

For training, we propagate the signal in ANN forward and measure the neural activity, \emph{i.e.} for all neurons $\alpha$ in the deep layers, indicated on the right panel of Figure~\ref{fig:experts}. 
We then compute the total neural activity and compare it with the neural activity in the desired expert network to calculate the loss function. 

The goal of the biomimetic Push-Pull loss function is to localize activity within the expert corresponding to the correct input class, while suppressing activity in all other experts. Therefore, the 'Push-Pull' loss function computes the difference between the logarithms of the total activity for each training set $m$, everywhere in the deep layers, and the activity in the correct expert: 
\begin{equation}
    \mathcal{L}_{PP} =\sum_{m=1}^M \left[ \log \left(\mbox{Total neural activity} \right)_m - 
    \log\left(\mbox{Activity in correct expert} \right)_m \right] \, .
    \label{loss_def_0}
\end{equation}
For a more precise definition of the loss function, see expressions \eqref{loss_def} in the \emph{Methods} section. This loss function is strictly non-negative, with $\mathcal{L}_{PP} =0$ achieved only in the highly unrealistic case when there are absolutely no excitations in the wrong experts for any training data. Notably, $\mathcal{L}_{PP}$ is structurally similar to conventional CCE but directly optimizes total metabolic activity rather than the final output logits.

Our Push-Pull formulation explicitly penalizes the spatial distribution of neural activity inside the hidden layers. Optimization of our loss function "pushes" total network activity down in the wrong experts while "pulling" the correct expert's activity up. As we demonstrate below, this mechanism not only matches or exceeds standard classification approaches but is uniquely capable of forcing the highly sparse, localized activation patterns necessary for modular, verifiable AI.

%This expert isolation is achieved through one of two approaches. In the first, \textit{structural segregation}, inter-expert connections are physically removed prior to training, so that each expert pathway is isolated by architectural design and cross-expert interference is eliminated by construction. In the second, \textit{functional segregation}, the full all-to-all connectivity is retained, but the Push-Pull loss encourages the network to develop strong lateral inhibition between experts during training, achieving approximate isolation through learned inhibitory weights rather than structural constraints.

As we mentioned,  we consider two different cases of biomimetic training using the Push-Pull loss function \eqref{loss_def_0}. The first case, corresponding to fully independent pathways and trained using the loss function \eqref{loss_def_0}, will be called the \emph{independent experts} case. In that case, inter-expert connections are eliminated by design.
In the second case, called the \emph{dense hybrid}, the model uses the full connectivity of the hidden layer and is trained with the Push-Pull loss function \eqref{loss_def_0}. That loss function encourages the network to develop strong lateral inhibition between experts during training, achieving a different type of expert isolation through learned inhibitory weights rather than structural design.  

Finally, to contrast our work with established approaches in image classification, we also use a fully dense network with the categorical cross-entropy (CCE), which is an established loss function for classification purposes.  Again, that loss function is applied to ANN training, with the accuracy verified by SNN,  as is described in details in \emph{Methods} section, equation \eqref{loss_CE}. 

There is a crucial difference between our method, which uses the expert architecture shown in Figure~\ref{fig:experts} with the loss function defined in \eqref{loss_def_0}, and standard methods such as CCE. Our loss function \eqref{loss_def} is explicitly designed to minimize neural activity in all the wrong experts and to enhance neural activity in the correct buckets, \emph{across all layers simultaneously}. In contrast, categorical cross-entropy \eqref{loss_CE} does not pose any restrictions on the signal propagation in the network, thereby not leading to sparse network structure and formation of isolated neural pathways. 

From a neurobiological perspective, the optimization of $\mathcal{L}_{PP}$
defined in \eqref{loss_def_0} and \eqref{loss_def} admits two distinct
biological interpretations. In the independent experts' case, the learning
dynamics mirror activity-dependent synaptic pruning during early neural
development \cite{katz1996synaptic,lichtman2000synapse}: the Push-Pull loss
carves out isolated, class-specific pathways by suppressing activity in
incorrect experts, replicating the ``use it or lose it'' principle of
competitive circuit formation. In the dense hybrid case, the analogy shifts
to REM sleep: the Push-Pull loss drives a functional shift in the
excitatory-inhibitory balance toward inhibition without structural elimination
of connections \cite{Yang2014,Zhou2020,Tamaki2020}, sharpening representational
separation through learned suppression rather than complete synaptic removal.

%From a neurobiological perspective, the optimization of $\mathcal{L}_{PP}$ defined in \eqref{loss_def_0} above, and, in more details, \eqref{loss_def} below, \textcolor{red}{in the independent experts case} closely mirrors the activity-dependent structural plasticity observed during early neural development \cite{katz1996synaptic}. In biological sensory systems, the formation of specialized functional circuits, such as discrete neuromeres in insects or ocular dominance columns in the mammalian cortex, begins with a dense overproduction of synapses, followed by a rigorous phase of competitive pruning \cite{lichtman2000synapse}. The dual nature of our objective function mathematically formalizes this developmental competition. The targeted enhancement of activity in the correct expert serves as an analogue to Hebbian long-term potentiation and localized neurotrophic reinforcement, where active, task-relevant synapses are stabilized \cite{poo2001neurotrophins}. Conversely, the global suppression of total internal activity imposes a simulated metabolic constraint; it drives heterosynaptic depression \cite{chistiakova2014heterosynaptic}, actively starving and pruning erroneous connections that project into the wrong expert domains. By enforcing this biological principle of ``use it or lose it'' at the architectural level, the learning process forces the network to physically decompose into highly isolated neural pathways.

\paragraph{Computational efficiency eliminating the need of SNN learning}
A major bottleneck in deploying spiking architectures is the computational overhead of training of direct methods such as Backpropagation Through Time (BPTT). To circumvent this, we optimized our decomposable architecture D-SNN for learning entirely within the continuous ANN using our differentiable surrogate, and subsequently ported the trained weights directly into the discrete SNN environment. The classification accuracy of the resulting network was verified entirely through pure SNN simulation.

Crucially, this direct-porting strategy yielded no significant degradation in performance while drastically reducing the computational burden during learning. Because the network avoids explicit SNN training dynamics with a temporal component, the learning procedure converges within minutes on standard consumer hardware (e.g., a laptop). In contrast, training an equivalent architecture via full BPTT typically requires one to two orders of magnitude more computational effort and necessitates high-performance computing clusters. This demonstrates that the D-SNN framework is both parameter-efficient at inference and computationally efficient during the training phase.

\subsection{Experimental results and comparison with the state-of-the-art: MNIST and Fashion-MNIST datasets}
\label{sec:MNIST}

To validate the classification capabilities and structural decomposition of the D-SNN, we benchmarked our architecture on the MNIST dataset of handwritten digits \cite{lecun1998gradient} and the more complex Fashion-MNIST (F-MNIST) dataset \cite{xiao2017fashion}, both of which serve as foundational benchmarks for assessing accuracy and parameter efficiency in the spiking neural network literature. 

For MNIST data set, we used a system with one layer of $F=8$ convolutional filters, and each of the experts in the hidden layers having $N_1=64$ neurons in the first layer and $N_2 = 32$ neurons in the second layer. We observed that the transition from a monolithic dense structure to isolated experts (One-vs-All, or OvA) does not incur a significant accuracy penalty. Our D-SNN  with Independent Experts and Dense Hybrid, for the example presented, both achieved an accuracy of 98.15\%, which is statistically indistinguishable from the 98.19\% achieved by standard dense networks trained with cross-entropy (Table~\ref{tab:snn_comparison}).
\begin{table}[hbt!]
\centering
\label{tab:snn_comparison_results}
\resizebox{\columnwidth}{!}{%
\begin{tabular}{@{}l l l r r@{}}
\toprule
\textbf{Method} & \textbf{Learning Rule} & \textbf{Architecture} & \textbf{Params} & \textbf{Acc} \\ \midrule
\textit{Unsupervised} & & & & \\
Diehl \& Cook (2015) \cite{diehl2015unsupervised} & STDP &  Dense$\times$2 & $\approx$  5 M & $95.0\%$ \\
Kheradpisheh et al. (2018) \cite{kheradpisheh2018stdp} & STDP + SVM &  Conv$\times$2 & $\approx$  80 K & $98.4\%$ \\ \midrule
\textit{Supervised} & & & & \\
Lee et al. (2016) \cite{lee2016training} & Backprop &  Conv$\times$2 + Dense$\times$2 & $\approx$  3.3 M &  99.3\% \\
Wu et al. (2018) \cite{wu2018spatio} & STBP &  Conv$\times$2 + Dense$\times$2 & $\approx$  3.9 M & $99.4\%$ \\ \midrule
%\textbf{Dense CCE} & \textbf{Cross-Entropy} \eqref{loss_CE} & \textbf{Conv-FC} & \textbf{$\approx$ 5.2 M} & \textbf{98.62\%} \\
\textbf{This work} & \textbf{Biomimetic  loss} \eqref{loss_def_0} & \textbf{Indep Experts} & \textbf{$\approx$ 272 K} & \textbf{98.2\%} \\ 
 & \textbf{Biomimetic  loss} \eqref{loss_def_0} & \textbf{Dense Hybrid} & \textbf{$\approx$ 460 K} & \textbf{98.2\%} \\ \bottomrule
\end{tabular}%
}

%\textcolor{red}{I would also list dense hybrid in the table}

%\textcolor{red}{table refs in the text (2.2),(2.3) does not match table labels (Table 1,2,3)}

\caption{  Benchmarking our architecture against standard SNN results on the MNIST dataset.  Conv and Dense refer to convolutional and fully-connected layers, respectively, and the symbol $\times$ indicates the number of such layers.  The independent experts topology, optimized via the biomimetic Push-Pull loss \eqref{loss_def_0}, achieves an SNN-verified accuracy of 98.15\%. Crucially, it delivers classification performance on par with highly entangled Dense CCE baseline (98.55\%) while utilizing a half of the trainable parameters ($\approx 272$K vs. $\approx 460$K). The savings of the parameter number achieved by physically severing lateral synaptic connections, demonstrates the computational advantages of structurally isolated processing pathways over dense global integration. Our architecture also allows for a much faster training procedure due to ANN to SNN conversion compared with the standard methods involving the full time dependence.  }
\label{tab:snn_comparison}
\end{table}

Remarkably,  in D-SNN with Independent Experts this accuracy parity was maintained despite a two-fold reduction in the number of trainable parameters within the hidden layers compared with the dense-hybrid and CCE methods. Our  'Push-Pull' training regime successfully condensed the necessary information into streamlined, specialist pathways. This demonstrates that for classification tasks, the vast majority of synaptic connections in a dense network are redundant, and can be pruned into isolated "ganglia" without compromising the system's global competence.

Fashion-MNIST presents a higher classification challenge compared to MNIST, requiring the network to distinguish between complex textures and shapes (e.g., T-shirt vs. Shirt vs. Coat). To address this, we scaled the architecture while maintaining the parallel expert design. The convolutional block capacity was increased up to 16 convolutional filters, and the expert paths were scaled to hidden layer sizes of $64$ and $32$ neurons per expert, so the whole network has the size $640$ in the first hidden layer and $320$ neurons in the second hidden layer. 
%The results of the learning procedure are presented on the left panel of Figure~\ref{fig:learning_FMNIST_CIFAR10}, with the particular case of number of filters being $F=16$. 
All architectures in our method achieved the accuracy of $\approx$91.7\% within about 100 epochs.   Notice that the independent experts case has approximately half the number of parameters compared with the other two cases, in spite of providing similar accuracy.

These results further demonstrate that the Push-Pull loss function and
structural segregation contribute independently to classification performance:
the Push-Pull loss alone, as implemented in the dense hybrid model, achieves
high accuracy by inducing functional modularity through learned inhibition,
while the additional structural isolation of the independent experts model
compounds these gains by eliminating residual crosstalk and reducing the
parameter count by an order of magnitude - an efficiency that learned
inhibition alone cannot achieve.

%%% END REM 
\begin{table}[hbt]
\centering
\resizebox{\columnwidth}{!}{% Resizes table to fit column width
\begin{tabular}{@{}l l l r r@{}}
\toprule
\textbf{Method} & \textbf{Learning Rule} & \textbf{Architecture} & \textbf{Params} & \textbf{Acc} \\ \midrule
\textit{Unsupervised} & & & & \\
Hao et al. (2020) \cite{hao2020biologically} & Reward STDP &  FC$\times$1 & $\approx$  5 M & 85.3\% \\
%Zhang et al. (2018) \cite{zhang2018plasticity} \\ \textcolor{blue}{[no data for FMNIST!]} & Eq. Prop. & FC-SNN (500-500) & $\approx$ 650 K & 86.6\% \\
%Kheradpisheh (2018) \cite{kheradpisheh2018stdp} \\ \textcolor{blue}{[no data for FMNIST!]} & STDP + SVM & Deep CNN (3-Lay) & $\approx$ 150 K & 87.3\% \\ 
\midrule
\textit{Supervised} & & & & \\
%Wu et al. (2018) \cite{wu2018spatio} \\ \textcolor{blue}{[no data for FMNIST!]}  & STBP & ConvNet (2-Conv) & $\approx$ 400 K & 89.8\% \\
Cheng et al. (2020) \cite{cheng2020lisnn} & LISNN &  Conv$\times$2 + Dense$\times$2 & $\approx$  2 M &  92.1\% \\
%Rathi et al. (2020) \cite{rathi2020diet} \\ \textcolor{blue}{[no data for FMNIST!]} & Diet-SNN & Deep VGG-16 & $\approx$ 15 M & 92.6\% \\ 
\midrule
\textbf{This work } & \textbf{Biomimetic loss} \eqref{loss_def_0} & \textbf{Indep Experts}  & \textbf{$\approx$ 351 K} & \textbf{91.7\%} \\ 
 & \textbf{Biomimetic  loss} \eqref{loss_def_0} & \textbf{Dense Hybrid}  & \textbf{$\approx$ 539 K} & \textbf{91.7\%} \\ \bottomrule
\end{tabular}
}
%\textcolor{red}{I would also list dense hybrid in the table}

\caption{ Comparison of the proposed decomposable Spiking Neural Network (D-SNN) against state-of-the-art unsupervised and supervised spiking frameworks on the F-MNIST dataset. %While traditional supervised methods rely on either massive parameter counts (e.g., Diet-SNN \cite{rathi2020diet}) or computationally expensive spatio-temporal backpropagation (STBP, \cite{wu2018spatio}), 
The D-SNN achieves competitive accuracy ($91.68\%$) using a modular, independent expert topology. Notably, our architecture maintains a sparse $\approx 350$k parameter footprint, while bypassing the computational overhead of including the temporal dynamics through biomimetic objective optimization \eqref{loss_def_0}.
}
\label{tab:fmnist_comparison}
\end{table}

As shown in Table~\ref{tab:fmnist_comparison}, while recent supervised SNNs exhibit strong classification performance on Fashion-MNIST, a direct comparison reveals a critical trade-off between inference efficiency and training complexity. Methods such as Spatio-Temporal Backpropagation (STBP) \cite{wu2018spatio} and LISNN \cite{cheng2020lisnn} achieve excellent parameter efficiency and competitive accuracy (89.8\% and 91.4\%, respectively). However, while these networks are compact, they achieve the accuracy at a high computational cost during training. These architectures rely on STBP, requiring the network dynamics to be expanded and investigated over time, which drastically increases memory overhead and training duration. Conversely, architectures like Diet-SNN \cite{rathi2020diet} achieve slightly higher accuracy (92.6\%) but require massive, parameter-rich network topologies (e.g., VGG-16 with $\simeq 15$ million parameters), making them unsuitable for cases with severe resource constraints.

Our decomposable architecture navigates this trade-off between the computational cost, number of parameters and accuracy. By achieving 91.68\% accuracy with  $\simeq 351$k parameters, the D-SNN suggested here remains highly competitive with state-of-the-art models. More importantly, it achieves this result without ever relying on temporal backpropagation. Because the independent experts are optimized entirely within the continuous ANN domain via our 'Push-Pull' loss function \eqref{loss_def_0} and subsequently ported directly to the discrete SNN environment, our method circumvents the massive training overhead of STBP and LISNN while entirely avoiding the large number of parameters of Diet-SNN.

To situate these results within the broader landscape of neuromorphic computing, we benchmarked our system against a state-of-the-art Backpropagation Through Time (BPTT) implementation. The BPTT baseline utilized an identical dense architecture (approx. 539K parameters) and achieved a peak accuracy of 91.12\%, similar to our independent experts. However, the computational cost of reaching this ceiling differed by orders of magnitude. The dense BPTT model required significant temporal unrolling and gradient calculation through time, leading to a high computational cost, approximately one hour on a GPU. Our ANN-based training of the modular system was approximately 10–50$\times$ faster depending on the architecture when trained on a laptop. Additionally, the independent expert architecture achieved this result with a two-fold reduction in active parameters (approx. 359K), demonstrating that enforced structural modularity can achieve state-of-the-art accuracy while radically lowering the energy and time overhead typical of deep spiking networks. 

\subsection{Applications of our method to CIFAR-10 dataset}
\label{sec:cifar}
To evaluate the scalability of the D-SNN architecture on natural image data, we extended our benchmarks to the CIFAR-10 dataset \cite{krizhevsky2009learning}. Consisting of color images with complex intra-class variance and background noise, CIFAR-10 tests whether isolated expert pathways can successfully maintain discriminative features when the 'sensory input' grows in complexity. To accommodate this, the shared convolutional 'retina' layers were expanded to three layers repeating the architecture of \cite{cao2015spiking}. In contrast, to challenge our method, we have used very small networks with $N_1=N_2=5$ neurons per expert in the hidden layer. The architecture of the hidden layer remained the same, comprising of either the independent experts, dense hybrid (connected experts), or the whole dense network trained by categorical cross-entropy (CCE).

As summarized in Table~\ref{tab:CIFAR10_snn_comparison}, in spite of extreme simplicity,  the dense hybrid case  achieves 83.3\% accuracy, followed by dense CCE and  independent experts with 82.8\% and 82.7\% respectively,  which are statistically indistinguishable.
A complete training from initialization to final SNN verification requires only $\sim 10$ minutes on standard consumer hardware (Apple M2 Max laptop). This reinforces the core advantage of the independent expert architecture: our methods provide highly resilient, computationally economical classification that for neuromorphic computing that can be done on a local workstation. 

\begin{table}[hbt]
\centering
\resizebox{\columnwidth}{!}{% Resizes table to fit column width
\begin{tabular}{@{}l l l r r@{}}
\toprule
\textbf{Method} & \textbf{Learning Rule} & \textbf{Architecture} & \textbf{Params} & \textbf{Acc} \\ \midrule
\textit{Standard SNNs} & & & & \\
Cao et al. (2015) \cite{cao2015spiking} & Conversion &  Conv$\times$3 + Dense$\times$2 & $\approx$ 150 K & 77.4\% \\
Hunsberger et al. (2015) \cite{hunsberger2015spiking} & SoftLIF &  ConvNet & $\approx$  1.5 M & 82.9\% \\
Wu et al. (2019) \cite{wu2019direct} & STBP & VGG-like CNN & $\approx$  28 M & 90.5\% \\
Rathi et al. (2020) \cite{rathi2020enabling} &  Conversion + STDB & VGG-16 & $\approx$ 15 M &  92\%\\ \midrule
\textit{This Work} & & & & \\
\textbf{D-SNN (F=64)} & \textbf{Biomimetic  loss} \eqref{loss_def_0}  & \textbf{Indep Experts} & \textbf{$\approx$ 350K} & \textbf{82.7\%}  \\ \textbf{D-SNN (F=64)} & \textbf{Biomimetic  loss} \eqref{loss_def_0}  & \textbf{Dense Hybrid} & \textbf{$\approx$ 352K} & \textbf{83.3\%}  \\ \bottomrule
\end{tabular}%
}
\caption{ 
Performance comparison between the proposed Independent Expert D-SNN and representative 
state-of-the-art spiking architectures on the CIFAR-10 dataset. Traditional supervised methods utilize deep, 
dense topologies such as ResNet-19 \cite{wu2019direct} and VGG-16 \cite{rathi2020enabling}, 
achieving high accuracy at the cost of massive parameter redundancy and intensive 
spatio-temporal backpropagation. In contrast, our D-SNN has only 350K parameters and is trained easily on a laptop without using computationally demanding iterations involving time-dependent dynamics used by all previous works in the table, except for \cite{cao2015spiking}. 
}

%\\ \\ \textcolor{blue}{[CIFAR-10 test does not look convincing because the accuracy gap is substantial. We need to have something comparable. For example, Cao et al. (2015) \cite{cao2015spiking} also seem to have very few parameters 0.3M and use ANN-to-SNN conversion, while achieving an accuracy of 77.4\% (actually, my calculations give 0.15M params for their architecture). Maybe we should tweak Conv layers in our D-SNN to achieve the same accuracy. That would look more competitive, but now it looks like D-SNN cannot solve CIFAR-10, and we are trying to find an excuse. For example, we can adopt their architecture completely \cite{cao2015spiking}: Conv1: 64 kernels 3x5x5 $\rightarrow$ MaxPool1 2x2 $\rightarrow$ Conv2: 64 kernels 64x5x5 $\rightarrow$ MaxPool2 2x2 $\rightarrow$ Conv3: 64 kernels 64x3x3 $\rightarrow$ FC layer1: 64 $\rightarrow$ Output 10. For independent experts, maybe we can do 10 experts with 10 neurons.]}}
\label{tab:CIFAR10_snn_comparison}

\end{table}

\subsection{Metabolic efficiency of the new architecture and learning procedure}
The biomimetic loss function \eqref{loss_def_0} enforces the suppression of extraneous neural activity while enhancing activity within the correct neural paths. It is therefore reasonable to expect substantially higher neural efficiency from the proposed architecture. To demonstrate this fact, we present results for \textit{Neural Sparsity}, defined as the average number of spikes in the hidden layers per neuron per unit time. These sparsity measurements are summarized in Table~\ref{tab:combined_sparsity_measurements}. Consequently, our architecture and learning procedure lead to substantial savings in energy, exceeding an order of magnitude even for the simplest classification tasks. We note that when the network dimensions are scaled up for the CIFAR-10 case, the firing rate within the independent experts drops substantially below 0.01 spikes/neuron/step. For brevity, we omit the presentation of these simulations from the main results, as our primary objective is to demonstrate the most compact network layout capable of achieving competitive classification performance.

One might naturally conjecture that our method yields even more substantial metabolic savings as the number of classes and the overall task complexity increase. This is indeed the case, as we verify in Section~\ref{sec:cifar-100} and illustrate in the right panel of Figure~\ref{fig:separation_CIFAR100}, demonstrating a sparsity improvement by a factor 3 in the dense hybrid model and of two orders of magnitude in the independent experts model.
\color{magenta} 
\begin{table}[hbt!]
\centering
\resizebox{\textwidth}{!}{%
\begin{tabular}{l ccc ccc ccc}
\toprule
\textbf{Architecture} & \multicolumn{3}{c}{\textbf{MNIST}} & \multicolumn{3}{c}{\textbf{Fashion-MNIST}} & \multicolumn{3}{c}{\textbf{CIFAR-10}} \\
\cmidrule(lr){2-4} \cmidrule(lr){5-7} \cmidrule(lr){8-10}
\textbf{Mode} & \textbf{Acc (\%)} & \textbf{FR} & \textbf{SynOps} & \textbf{Acc (\%)} & \textbf{FR} & \textbf{SynOps} & \textbf{Acc (\%)} & \textbf{FR} & \textbf{SynOps} \\
\midrule
Dense BPTT    & \textbf{98.56} &  0.095        & 2.04M & 91.12 & 0.174          & 4.47M & \textbf{83.49} & 0.361         & 150K \\
Dense CCE     & 98.18 & 0.247         & 2.56M & 91.67 & 0.377          & 6.02M & 82.82 & 0.427          & 131K \\
Dense Hybrid   & 98.15 &  0.044          &   492K & \textbf{91.68} & \textbf{0.035}          & 311K & 83.27 & \textbf{0.069} &  19.3K \\
Indep Experts  & 98.15 & \textbf{0.041} & \textbf{37K} & \textbf{91.68} &  0.045 & \textbf{36K} & 82.74 & 0.071          &  \textbf{1.89K} \\
\bottomrule
\end{tabular}%
}
\caption{\textbf{Classification performance, Firing Rate Efffiency, and Synaptic Operations across different network topologies.} Spiking neural network (SNN) verification accuracy, metabolic sparsity (measured as mean population firing rate per timestep, FR), and average backend synaptic operations per sample (SynOps) are compared across the MNIST, Fashion-MNIST, and CIFAR-10 datasets. Network capacities were standardized per dataset to ensure fair benchmarking: MNIST and Fashion-MNIST utilized a base filter size of $F=16$ with hidden layers $N_1=64$ and $N_2=32$, while the more complex CIFAR-10 task utilized a three-layer convolutional backbone with $F=64$ and a downscaled hidden topology of $N_1=5$, and $N_2=5$. SynOps calculations compute the neural activity in the hidden layers only (backend), from the first to the second hidden layer ($FC_1 \rightarrow FC_2$) and from the second hidden layer to the output layer ($FC_2 \rightarrow FC_3$) to isolate structural routing mechanics from static frontend expansions. Enforcing independent, block-diagonal connectivity combined with a biomimetic hybrid loss function \eqref{loss_def_0} reduces redundant spiking by up to an order of magnitude and slashes synaptic operations by several orders of magnitude %(a $\sim$400$\times$ reduction on MNIST, $\sim$ 2,00$\times$ reduction on Fashion-MNIST and $\sim$70$\times$ on CIFAR-10) 
with negligible impact on task accuracy. Finally, performance is evaluated against a baseline full BPTT calculation incorporating complete unrolled temporal dynamics, which tends to maximize classification accuracy, albeit at a significantly higher training cost. }
\label{tab:combined_sparsity_measurements}
\end{table}
\color{black} 
These results demonstrate that the Push-Pull loss alone substantially
increases sparsity over standard CCE training, confirming that learned
inhibition between expert groups actively suppresses spurious activity
across pathways. However, the dense hybrid consistently falls short of
the sparsity achieved by independent experts,
revealing a fundamental limitation of the inhibitory approach: learned
suppression can reduce but never fully eliminate cross-pathway activity
as long as the underlying connections remain intact. Structural segregation,
by contrast, removes the source of interference entirely, achieving a
level of metabolic efficiency that no amount of learned inhibition can
replicate.

To additionally measure the efficiency, we incorporate the commonly used metric of 
synaptic operations (SynOps) \cite{bu2025activity}. To explicitly isolate the structural efficiency of 
our routing mechanics, we monitor the backend activity where  our architecture is applied, \emph{i.e.} the asynaptic operations from the first to the second hidden layer, and from the second hidden layer to the output layer, since the architecture of the convolutional layers is identical across the architectures. 
This measure is used alongside 
the average firing rate to estimate the total energy consumption of the system. We note that our solution, which aggregates outputs over 
$T=100$ time steps, is intentionally chosen to be quite long; in practice, accurate 
classification can be achieved using a substantially shorter time window, further 
reducing the number of required synaptic operations.

For MNIST, the average number of backend synaptic operations per sample was 
approximately 2.56M for dense CCE, $\sim$500K for dense hybrid, and a mere 37K for 
independent experts. This demonstrates a massive $\sim$70-fold improvement in 
structural efficiency over the standard baseline, and a $\sim$37-fold reduction 
over the dense hybrid architecture alone.

For Fashion-MNIST, dense CCE requires an average of $\sim$6M synaptic operations per 
sample, dense hybrid requires 311K, and independent experts require only 36.2K 
operations, yielding a $\sim$160-fold improvement in efficiency. This result directly 
highlights the metabolic efficiency of our independent experts architecture and 
the efficacy of our specialized learning procedure using the loss function \eqref{loss_def_0}.

Furthermore, by strictly measuring the backend routing efficiency, we observe 
profound energy savings even with the exceptionally small hidden layer sizes 
utilized for the CIFAR-10 dataset. The average backend synaptic operations required 
for a classification decision plummet from roughly 130K per sample for dense CCE, 
down to 19.3K for dense hybrid, and to only 1.89K for independent experts. This 
confirms that the metabolic efficiency and the benefits of our block-diagonal 
structural sparsity are remarkably robust; even in highly downscaled topologies, 
our method achieves nearly a 69-fold reduction in compute cost without sacrificing 
task accuracy.

On Table~\ref{tab:combined_sparsity_measurements}, we also present the results of BPTT simulations for all problems (MNIST, Fashion-MNIST and CIFAR-10) which takes into account the temporal dimension of the system. Because of that factor, the training procedure using BPTT takes substantially more time, typically 10-50x more on the same hardware after all accelerations are taken into account. BPTT-trained networks retain the density and complexity and thus are not expected to have the energy efficiency of our architectures. 

\subsection{Internal organization of synaptic features: the FMNIST example}
\label{sec:internal_synaptic}

%\textcolor{blue}{[Are all the Figures 100\% about CIFAR-10? Because Fig 5 is for FMNIST]}

\begin{figure}
    \centering
    \includegraphics[width=1\linewidth]{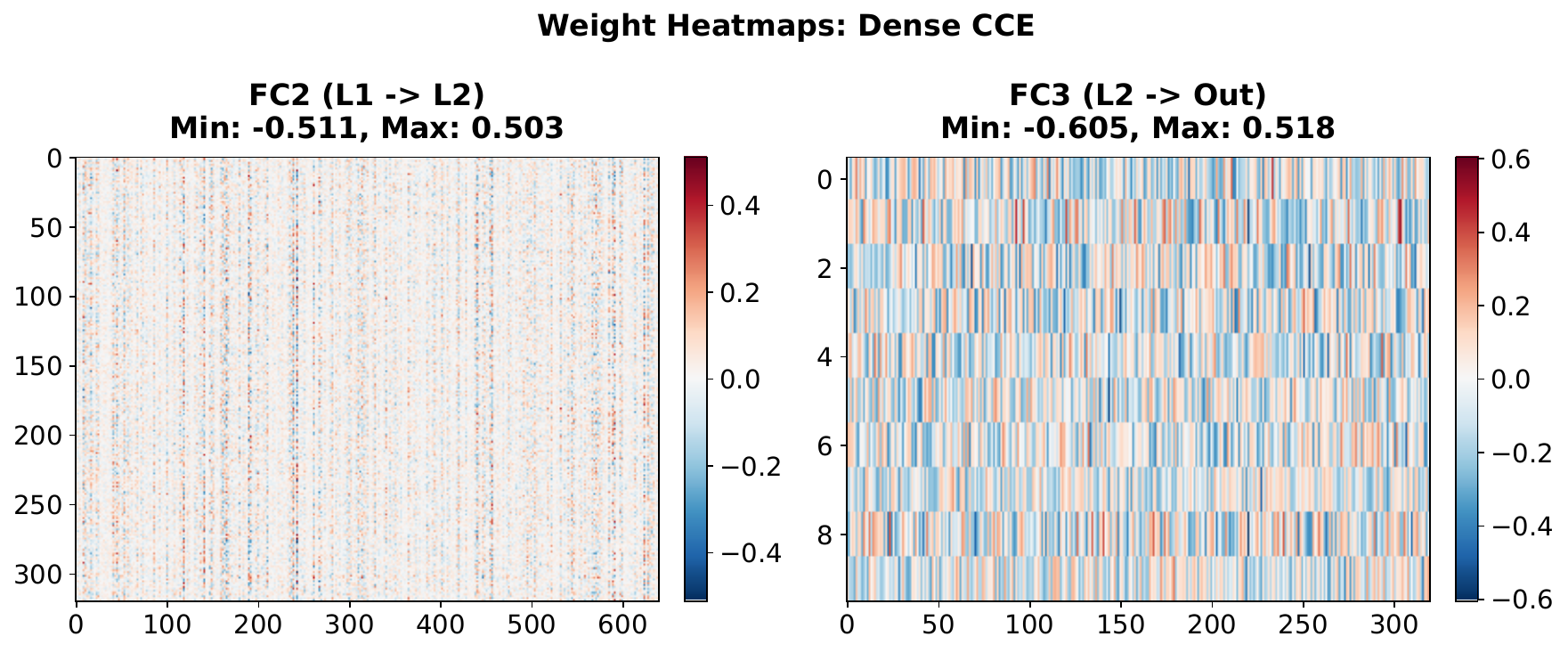}
     \includegraphics[width=1\linewidth]{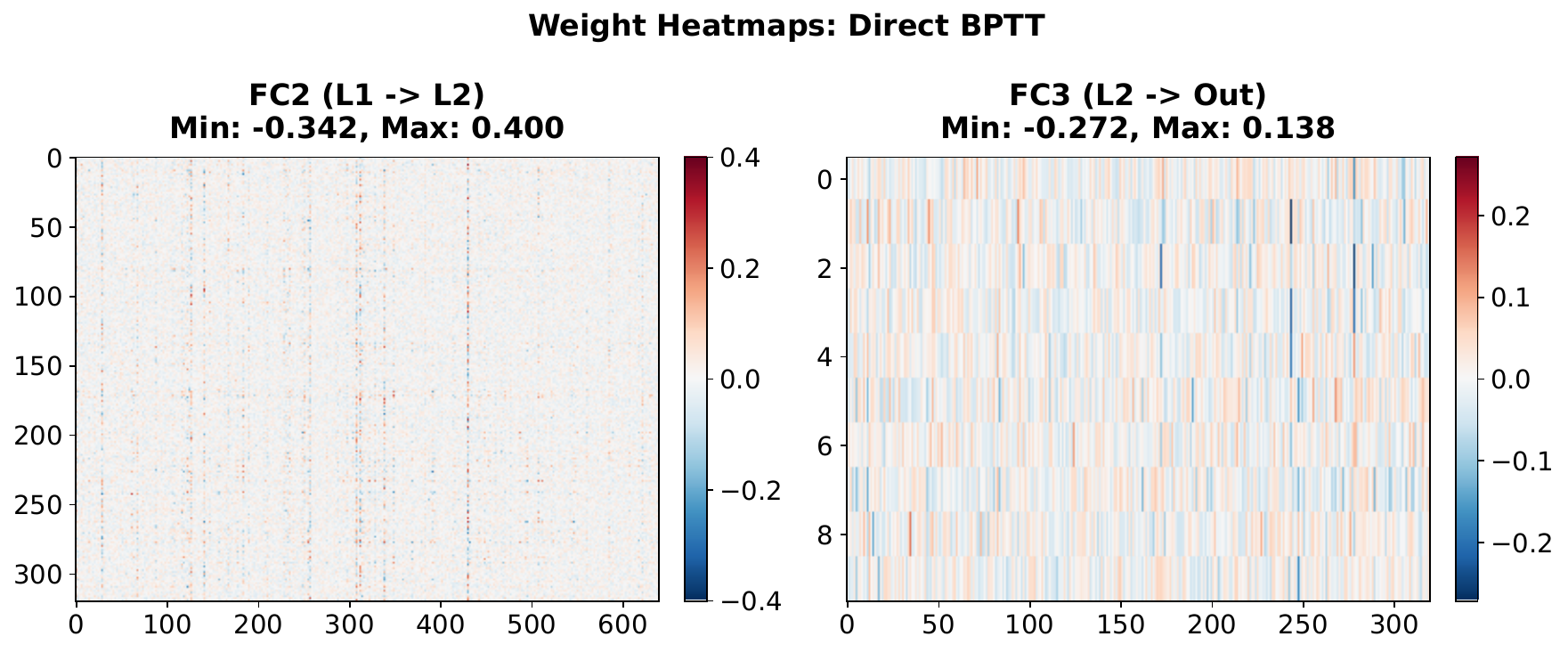}
 \caption{ Absence of structural modularity in SNN optimizated by standard global optimization methods on Fashion-MNIST dataset, 
using weight matrix visualizations for dense networks.  Top: Weights optimized via standard Cross-Entropy (CCE).  Bottom: Weights optimized via Spatio-Temporal Backpropagation Through Time (BPTT). For both methods, the panels display the synaptic weights from hidden layer 1 to hidden layer 2 (Left), from hidden layer 2 to the output layer (Right). In all cases, optimization results in a highly disordered, globally entangled weight distribution without any emergent class-specific isolation. This confirms that standard loss landscapes do not spontaneously generate modularity, establishing the baseline against which the decomposable architecture (Figure~\ref{fig:weights_dense_OvA_PP}) is compared.}
    \label{fig:weights_dense_CE_BPTT}
\end{figure}

\begin{figure}
    \centering
    \includegraphics[width=1\linewidth]{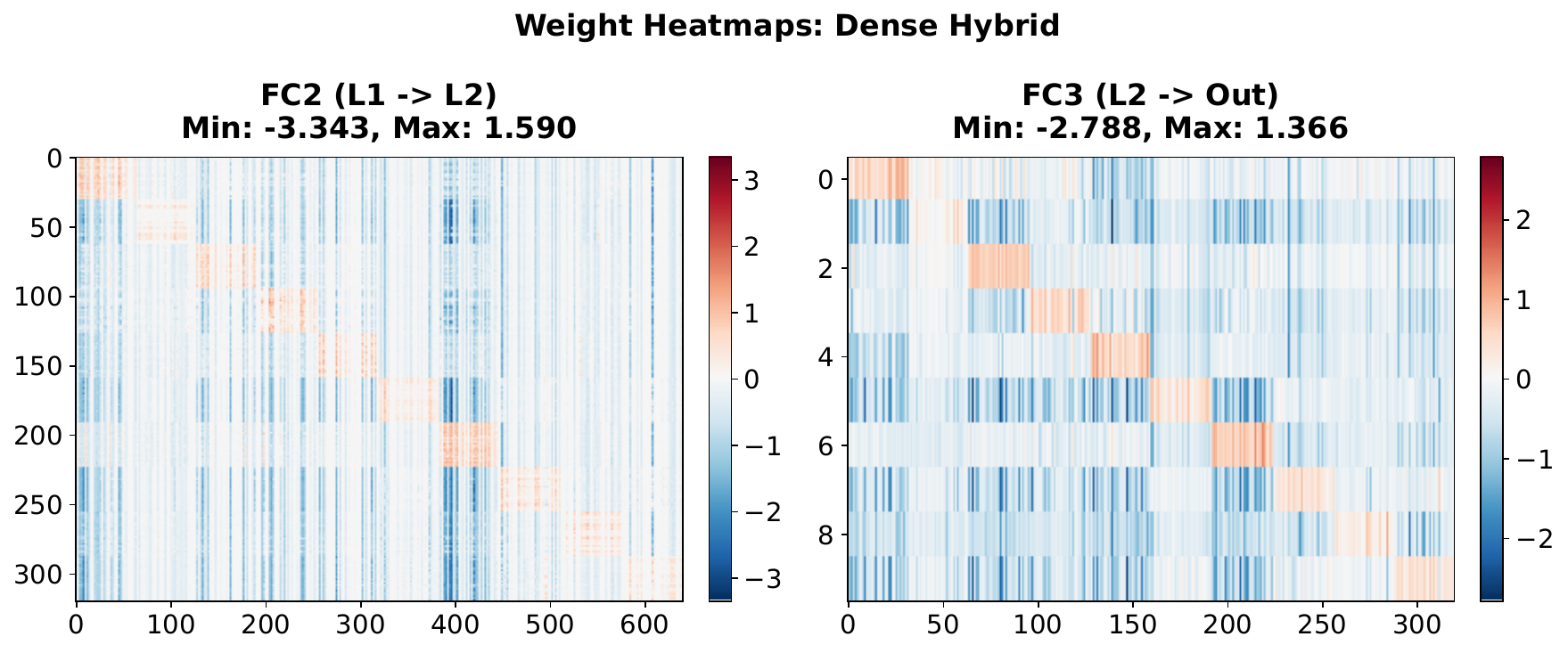}
     \includegraphics[width=1\linewidth]{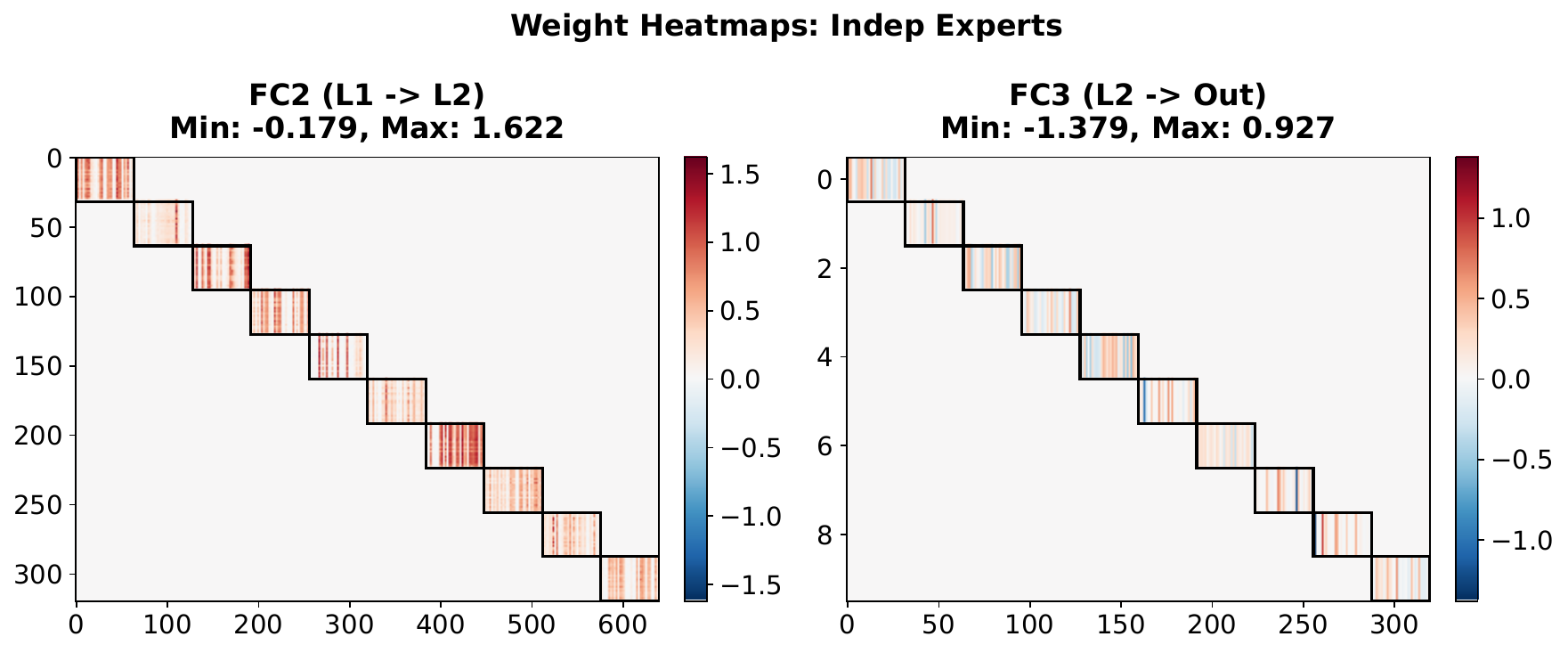}
 \caption{ Emergence of functional and structural modularity driven by biomimetic optimization on the Fashion-MNIST dataset.
Internal weight representations demonstrating the effect of the 'Push-Pull' objective \eqref{loss_def_0} 
on network topology. Top: Dense Hybrid. Application of the biomimetic loss to a fully 
connected architecture. Unlike the entangled weights of standard optimization, illustrated on Figure \ref{fig:weights_dense_CE_BPTT}, 
the competitive objective induces clear functional decoupling, evidenced by strong lateral 
inhibition (negative weights) where each emergent expert actively suppresses competing neural 
pathways.  Bottom: Independent Experts. The corresponding weight matrix for the strictly 
partitioned architecture. By physically decoupling the pathways, off-diagonal cross-talk 
parameters are reduced to absolute zero (indicated by solid black boundaries). This eliminates 
the need for dynamic inhibition, transforming the energetic metabolic competition of the 
dense hybrid into absolute, highly verifiable structural modularity.
 %\\ \textcolor{blue}{[Maybe after submitting, but I would massage the figure to make look nicer. What we can do (I can do it myself if I have the data for the plots): 1) sub-figures of exactly the same size (they are slightly different) due to colorbars]
 }
    \label{fig:weights_dense_OvA_PP}
\end{figure}

To investigate the internal organization of learned features, we visualized the synaptic weight matrices across different optimization and architectural regimes. In Figure \ref{fig:weights_dense_CE_BPTT}, we optimized dense networks via standard global objectives, such as Categorical Cross-Entropy (CCE) for the surrogate ANN training, and Back-Propagation Through Time (BPTT). The latter method, requiring incorporation of the temporal dependence, makes the training procedure particularly computationally intensive. Both cases exhibit a highly disordered weight distribution without any observed patterns, as expected. These standard approaches entangle features globally and do not generate any discernible class-specific isolation of neural pathways. In contrast, applying the minimization of our biomimetic 'Push-Pull' loss function \eqref{loss_def_0} fundamentally alters the network topology. When applied to a dense network (Fig. \ref{fig:weights_dense_OvA_PP}, Top), the competitive metabolic landscape induces emergent functional modularity; the network develops strong lateral inhibition (visible as dense regions of negative weights) to actively suppress competing class pathways. Finally, the proposed independent experts architecture, shown on the bottom of Figure~\ref{fig:weights_dense_OvA_PP}, has, by design, the absolute isolation of the experts from each other. The off-diagonal crosstalk parameters are enforced to be zero, completely eliminating the computational costs of active inhibition. 

\begin{figure}
    \centering
    \includegraphics[width=0.48\linewidth]{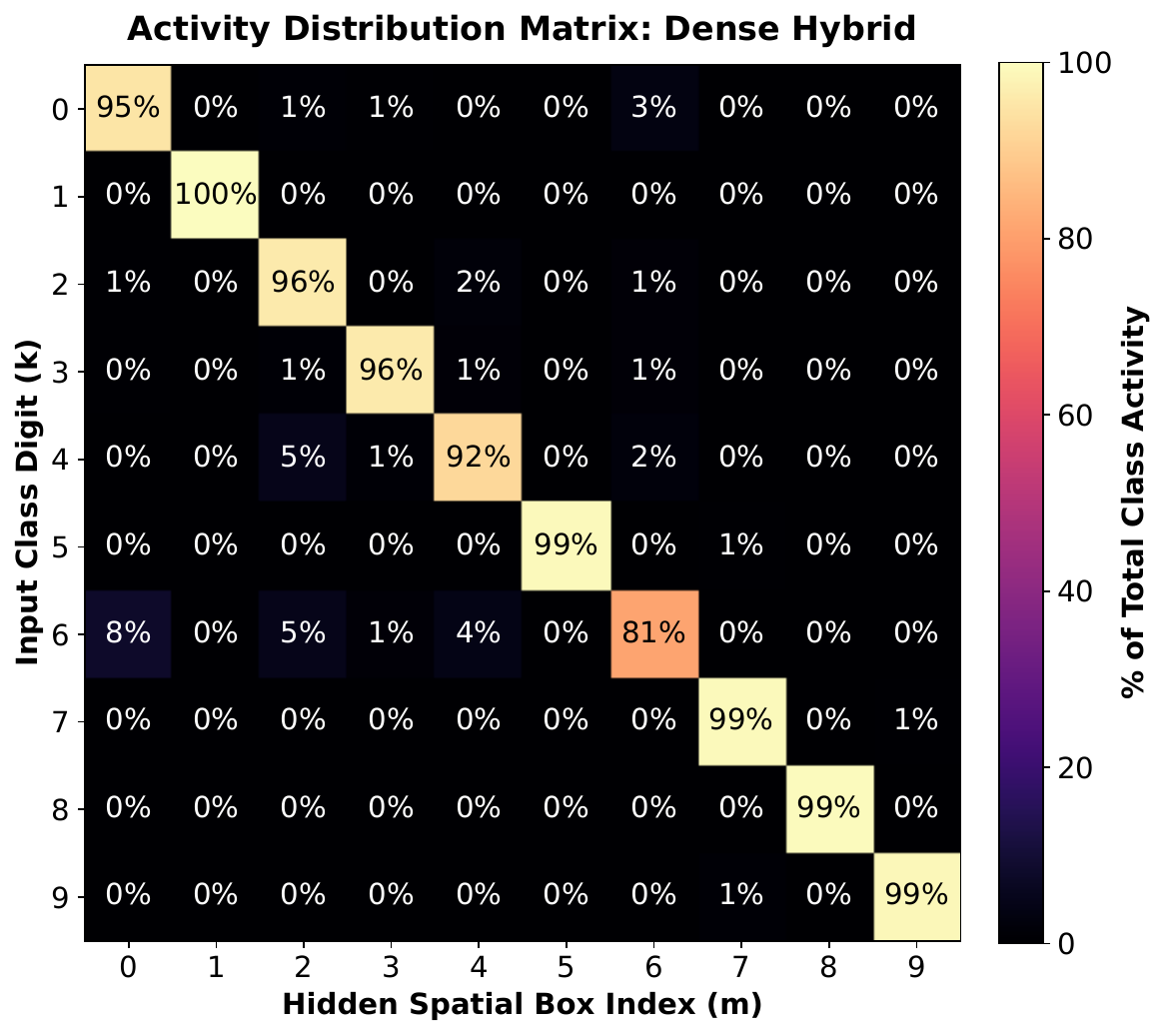}
    \includegraphics[width=0.48\linewidth]{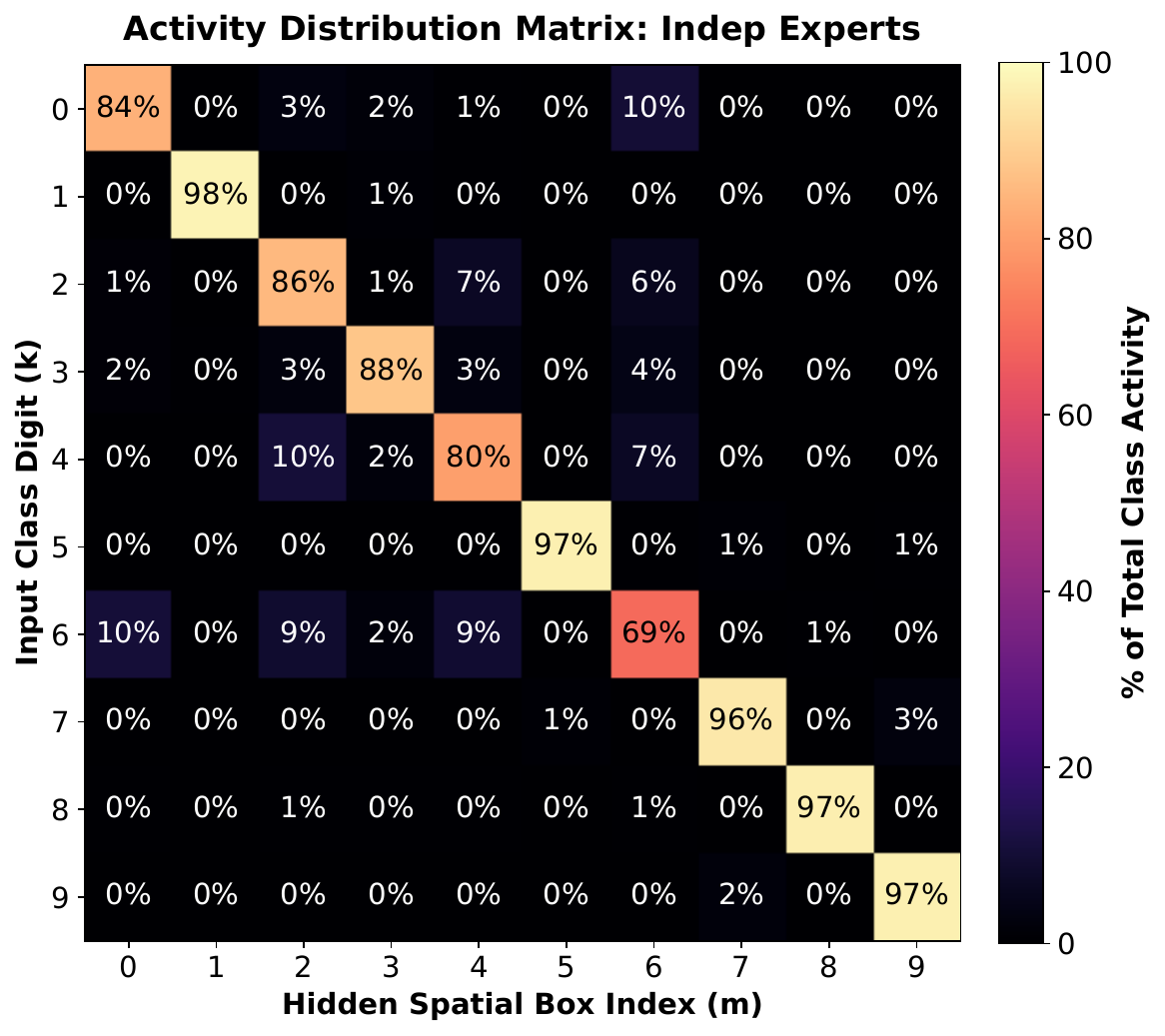}
    \caption{Comparative visualization of spiking activity localized within class-specific expert pathways during inference for Fashion-MNIST. Neural activity under the 
'Push-Pull' objective \eqref{loss_def_0} within a fully connected topology (Left, Dense Hybrid;  Right: Independent experts).  While the loss 
encourages functional localization, some minimal residual crosstalk persists, forcing the network to 
expend metabolic energy on active lateral inhibition. That cross-tallk, however, is negligible compared to dense architectures. Crucially, both architectures achieve equivalent peak classification accuracy 
($\approx 90\%$) and make an efficient use of the neural activity. We will further study the energy efficiency of this method in Section~\ref{sec:cifar-100}, leading to results illustrated on Figure~\ref{fig:separation_CIFAR100}. }
    \label{fig:activity_dense_OVA_PP}
\end{figure}

To evaluate the metabolic efficiency and traceability of these structural configurations, we analyzed the spatial distribution of spiking activity during inference (Fig. \ref{fig:activity_dense_OVA_PP}). In the dense hybrid model, although the biomimetic objective successfully biases activity toward the target class pathway, the underlying global connectivity makes the signal globally entangled and not representative of any particular class. In contrast, the independent experts architecture achieves almost perfect spatial sparsity. Because the pathways are structurally decoupled, sensory input routed to non-target experts naturally decays without triggering interference. During inference, the correct expert pathway fires actively while the remainder of the network remains quiescent. While both the dense hybrid and the independent experts achieve practically identical classification accuracy ($\approx 90\%$), the strict isolation in the latter guarantees that significantly fewer spikes are generated per decision. This "architectural silencing" provides a dual advantage: it offers a highly verifiable, traceable map of network decision-making, and opens a path to designing energy-efficient, sparsely activating networks that mimic biological organisms.

\subsection{Image Classification on CIFAR-100 with Limited Resources}
\label{sec:cifar-100}

In order to show the scalability of our method to substantially more complex problems compared to the classification of datasets with ten classes, we extended our analysis from the categorical boundaries of CIFAR-10 to the high-density, fine-grained landscape of CIFAR-100 \cite{krizhevsky2009learning}. This dataset provides a rigorous stress test, requiring the network to isolate features across highly overlapping classes under strict resource constraints. By varying the target subset size $M \in \{10, 20, \dots, 100\}$, we establish a continuous scaling benchmark bridging macroscopic classification with microscopic discrimination. Our goal is to investigate how our independent experts method performs when $M$ increases, with a specific focus on demonstrating the profound structural sparsity and corresponding energy savings it provides for increasingly complex problems.

For these scaling experiments, the shared convolutional front-end filter architecture was configured identically to the CIFAR-10 setup (three convolutional layers of 64 filters each). To maintain strict parameter parity across architectures, the downstream fully connected network was constrained to a fixed total width of $N=500$ neurons. The independent experts and dense hybrid networks distributed these hidden neurons evenly across the $M$ target classes, ensuring each isolated pathway contained an identical parameter budget per layer. 

\begin{figure}[hbt!]
    \centering
    \includegraphics[width=0.49\linewidth]{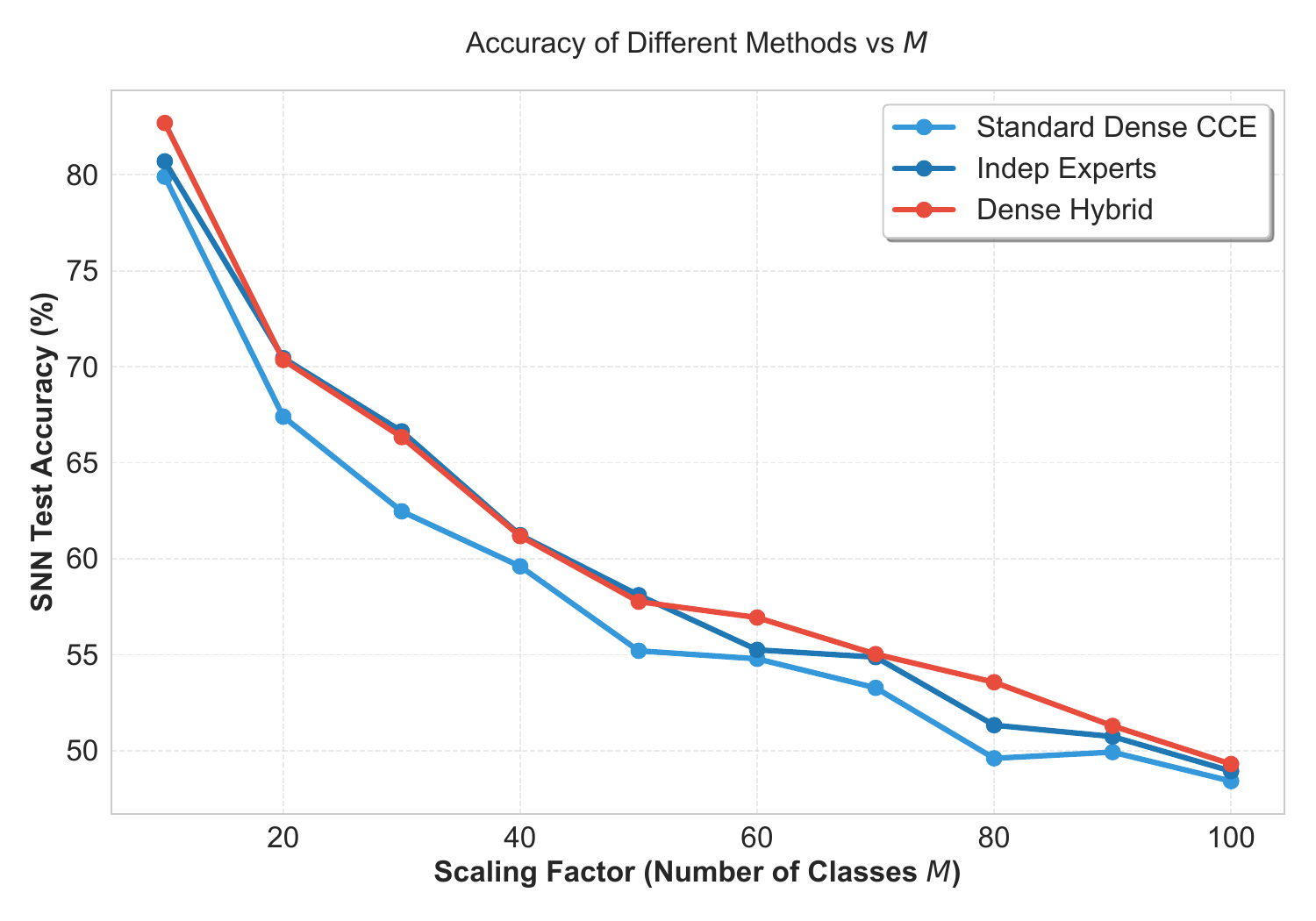}
    \includegraphics[width=0.49\linewidth]{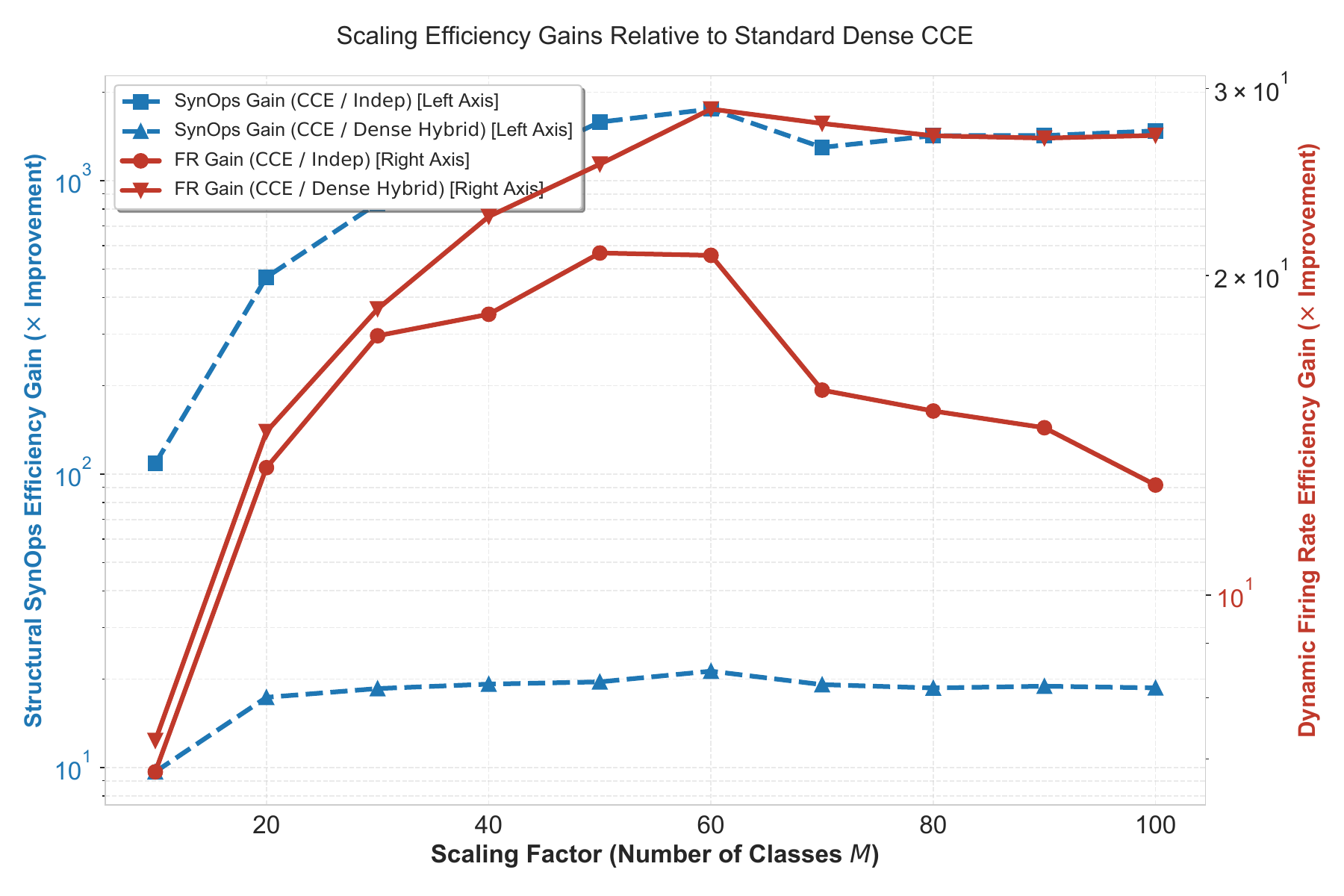}
    \caption{Left panel: Plot of classification accuracy as a function of the number of target classes ($M$) drawn from the CIFAR-100 dataset. All methods yield roughly similar, highly competitive classification capabilities as the problem scales, with the independent experts architecture consistently achieving slightly higher accuracy. Right panel: the  energy efficiency of our architectures as measured in the deep layers neural activity.  The right panel demonstrates the scaling efficiency multipliers relative to standard dense CCE. The dynamic firing rate efficiency (red lines, right axis) shows over an order of magnitude improvement for the hybrid configurations. Crucially, the structural routing efficiency (blue lines, left axis) demonstrates that the independent experts architecture completely bypasses the quadratic scaling penalty of dense networks, achieving several orders of magnitude ($>$1000$\times$) fewer backend synaptic operations at $M=100$.
    % \\ \textcolor{blue}{[Maybe after submitting, but I would massage the figure to make look nicer. What we can do (I can do it myself if I have the data for the plots): 1) font sizes of labels and ticks are larger and the same for both sub-figures.]}
    }
    \label{fig:separation_CIFAR100}
\end{figure}

As the number of classes $M$ increases, the SNN-computed accuracy demonstrates that the accuracy achieved by each method is similar across all three architectures,  as shown in the left panel of Figure~\ref{fig:separation_CIFAR100}. This Figure represent the best result across ten independent trials with the same parameters for each architecture, and for each value of $M$. The architectures developed using our loss function \eqref{loss_def_0} (dense hybrid and independent experts) consistently achieve slightly higher accuracy than the standard dense CCE model across the entire scaling continuum. 

While classification capabilities remain roughly equivalent, the true difference between the architectures emerges in their energy footprints, as illustrated in the right panel of Figure~\ref{fig:separation_CIFAR100}. To quantify this, we plot the efficiency multipliers, calculated as the ratio of the metabolic and structural costs of standard dense CCE against our proposed methods. The red curves (right axis) demonstrate the dynamic firing rate (FR) efficiency; driven by the biomimetic push-pull loss function \eqref{loss_def_0}, both hybrid configurations successfully suppress redundant spiking, yielding over an order of magnitude ($\sim$10$\times$) improvement in the firing rate. 

The energy efficiency of the independent expert architecture due to large savings in synaptic operations, compared to the fully dense models, is demonstrated on the right panel of Figure~\ref{fig:separation_CIFAR100} with blue curves, related to the left axis. That efficiency can be estimated analytically as follows. In a dense network, backend routing connections scale quadratically $O(M^2)$ due to the global fan-out of every spike. Conversely, the block-diagonal structure of the independent experts restricts spike propagation exclusively to the $1/M$ fraction of downstream neurons in its designated pathway, yielding the efficiency roughly proportional to linear $O(M)$ scaling.  Because of this architectural decoupling, the SynOps efficiency multiplier for the independent experts grows with $M$. By the extreme limit of $M=100$, the independent experts architecture computes inferences using three orders of magnitude fewer backend synaptic operations compared to the dense CCE baseline. 

\begin{figure}[!ht]
    \centering
    \includegraphics[width=0.49\linewidth]{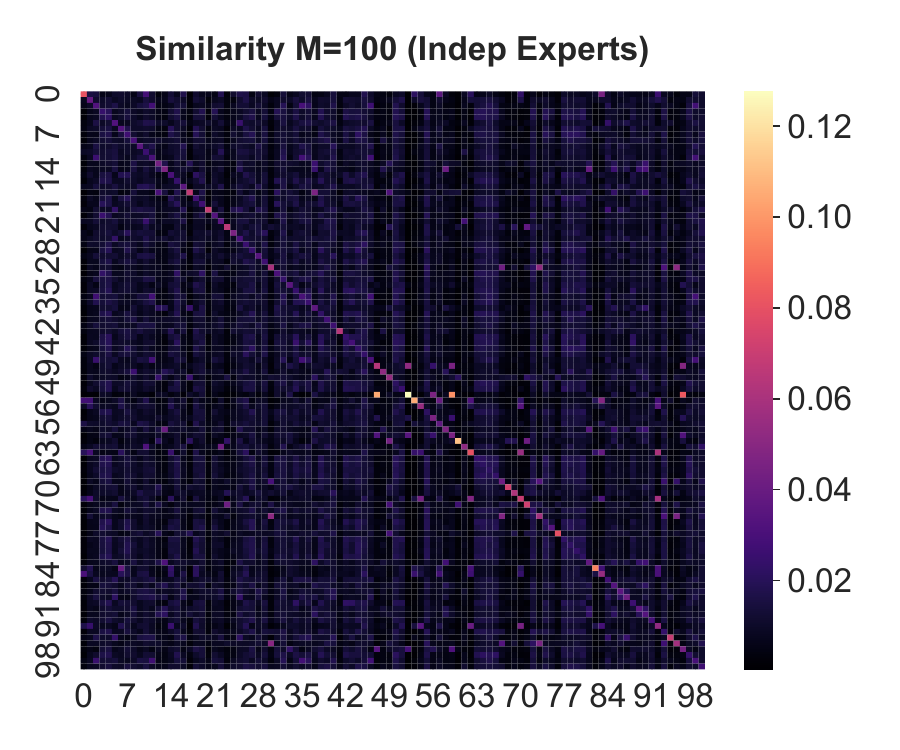}
    \includegraphics[width=0.49\linewidth]{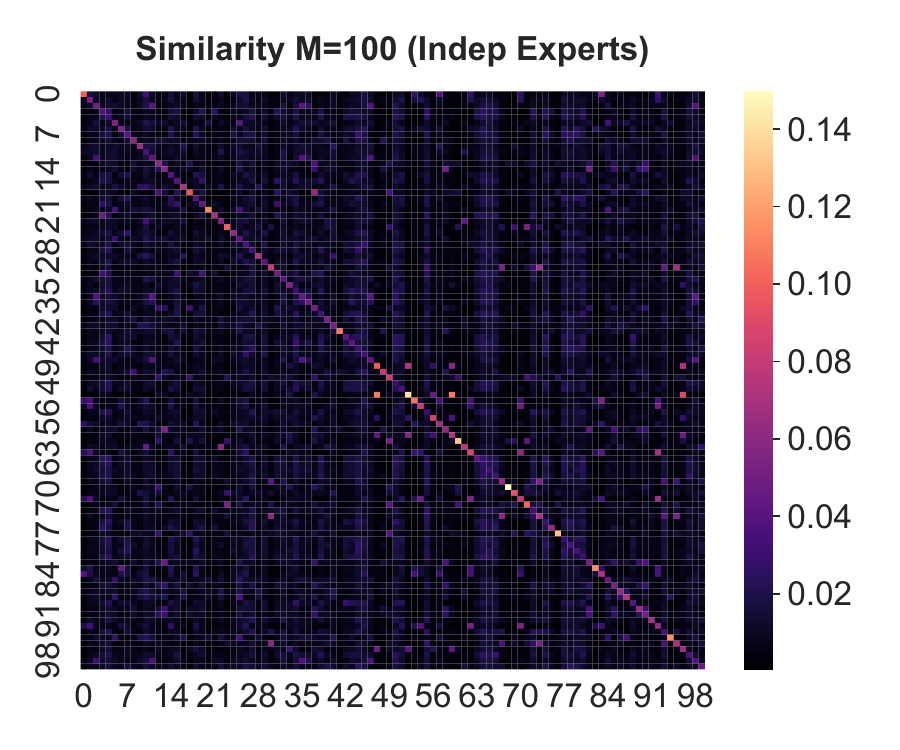}\\
    \includegraphics[width=0.49\linewidth]{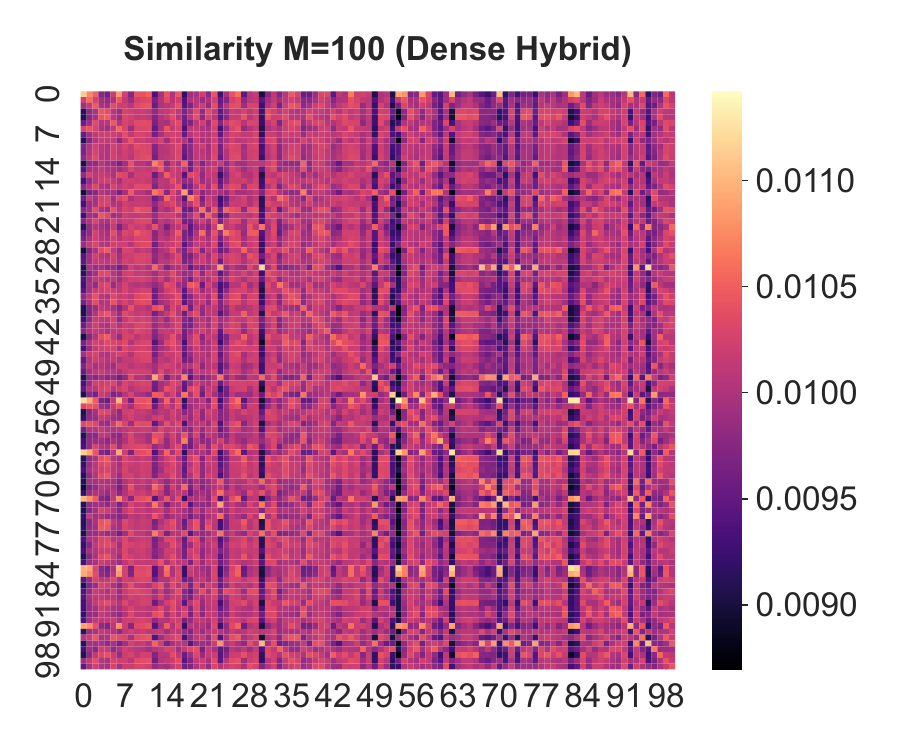}
    \includegraphics[width=0.49\linewidth]{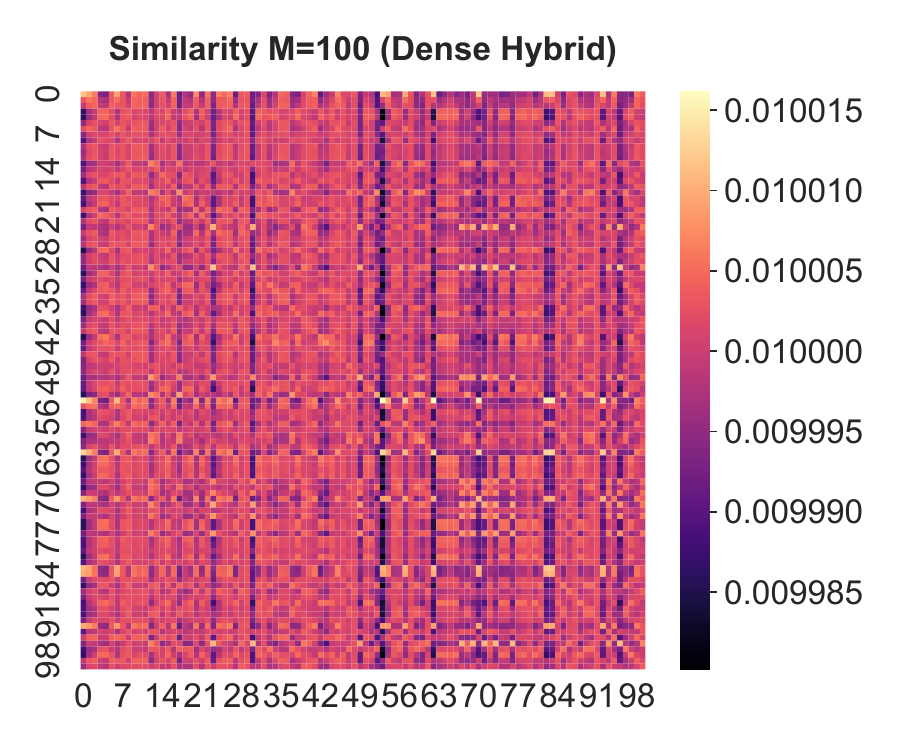}
    \caption{ Orthogonalization of spiking activity and signal fidelity at maximum classification difficulty ($M=100$), shown through spatio-temporal activity metrics for the independent experts architecture on the CIFAR-100 benchmark. In this highly constrained configuration, each of the $100$ independent expert pathways operates with an extremely limited capacity of only $5$ neurons per hidden layer. Top row: Matrices displaying the average spike counts in hidden layer 1 (left panel) and hidden layer 2 (right panel). Element $(k,j)$ represents the mean activity in expert $j$ when presented with ground-truth class $k$. Thus, the horizontal axis represents the specific expert pathway being monitored (labeled by its target class index), while the vertical axis represents the actual ground-truth stimulus class presented to the network. The dominant diagonal structure indicates that neural input is successfully routed exclusively to the target expert with minimal leakage to non-target pathways.
    %Bottom row: Correlation matrices of activation vectors between classes for hidden layer 1 (\textbf{Left}) and hidden layer 2 (\textbf{Right}). Despite the severe reduction in parameter volume and the high morphological overlap of CIFAR-100 classes, the structurally isolated pathways maintain highly orthogonal feature representations. This prevents the catastrophic feature entanglement that drives capacity collapse in monolithic topologies.
    % \\ \textcolor{blue}{[Maybe after submitting, but I would massage the figure to make look nicer. What we can do (I can do it myself if I have the data for the plots): 1) 1 colorbar per row of sub-figures; 2) colorbars can have a multiplier on top, like 1e-3 so that the ticks are more compact; 3) same-sized sub-figures for easier reading; 4) shown X and Y axes ticks only for (0, 10, 20, ..., 99); 5) Y ticks are rotated to be horizontal.]}
    }
    \label{fig:separation_activity_M100}
\end{figure}

To uncover the mechanism underlying this massive efficiency, we analyzed the class-conditional spiking distributions at the absolute limit of our scaling benchmark ($M=100$). Figure~\ref{fig:separation_activity_M100} visualizes the correlations of the spiking signal during the internal neural dynamics of this highly constrained network. The mean spike count matrices reveal a pronounced diagonal structure across both computational stages. This confirms that despite the complex, overlapping visual space of CIFAR-100, class-specific stimuli are successfully routed exclusively to their correct specialist pathways. Because non-target experts remain metabolically dormant due to the optimization procedure given by \eqref{loss_def_0}, the activation vectors for different classes are essentially orthogonal. This structural isolation ensures that the network is perfectly auditable at any point in time and fundamentally prevents the inefficient energy fan-out inherent to fully connected networks.

\subsection{Sparse architecture prevents catastrophic forgetting in continual learning
}

%Humans and other animals can learn continuously from few examples, integrate new data into existing knowledge, and generalize beyond single experiences. In contrast, artificial neural networks (ANNs) suffer “catastrophic forgetting,” performing well on new tasks at the expense of earlier ones \cite{french1999catastrophic,hayes2021replay,mcclelland1995there,mccloskey1989catastrophic}.

Humans and other biological organisms possess the capacity for continual learning; they can integrate new information from limited examples without overwriting existing knowledge \cite{french1999catastrophic,hayes2021replay}. In contrast, standard artificial neural networks (ANNs) suffer from catastrophic forgetting, where adapting to new tasks severely degrades performance on previously learned ones \cite{mcclelland1995there,mccloskey1989catastrophic}.

We evaluated the D-SNN under a continual learning paradigm in which the ten
MNIST classes were split into two sequential tasks: Task~1 comprised classes
$0$ through $M$ and Task~2 classes $M+1$ through $9$, for $M \in \{1,
\ldots, 8\}$. The networks were first trained on Task~1, after which Task~2
was introduced alongside a severely constrained rehearsal dataset representing
Task~1.  To emulate biological learning, the network was not given access to the
original Task~1 training images during Task~2 learning. Instead, the
rehearsal dataset consisted of synthetic images, each formed by averaging
all Task~1 training images of a given class to produce $M$ class templates,
which were then augmented with random Gaussian noise, following earlier
works on sleep and memory consolidation \cite{Tadros2020Biologically,Kubo2025}.

Because dense networks rely on globally integrated representations, accommodating 
the new classes from this severely restricted data forces gradient updates across the entire 
synaptic matrix, overwriting established weights. 

Conversely, for the independent experts and dense hybrid architectures, it is straightforward to preserve memory by manually freezing the weights corresponding to the experts in the old classes (which is impossible in dense CCE networks).

\begin{figure}[tbh!]
    \centering
        \includegraphics[width=0.49\linewidth]{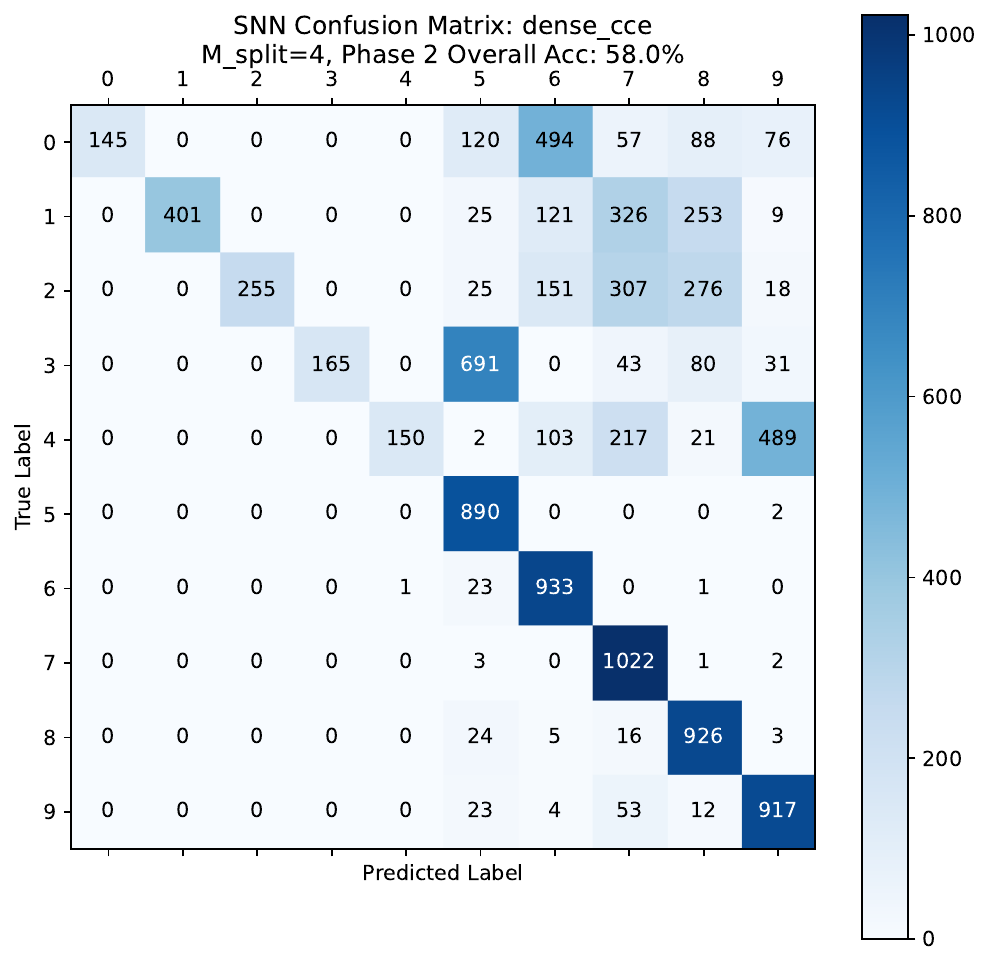}
        \includegraphics[width=0.49\linewidth]{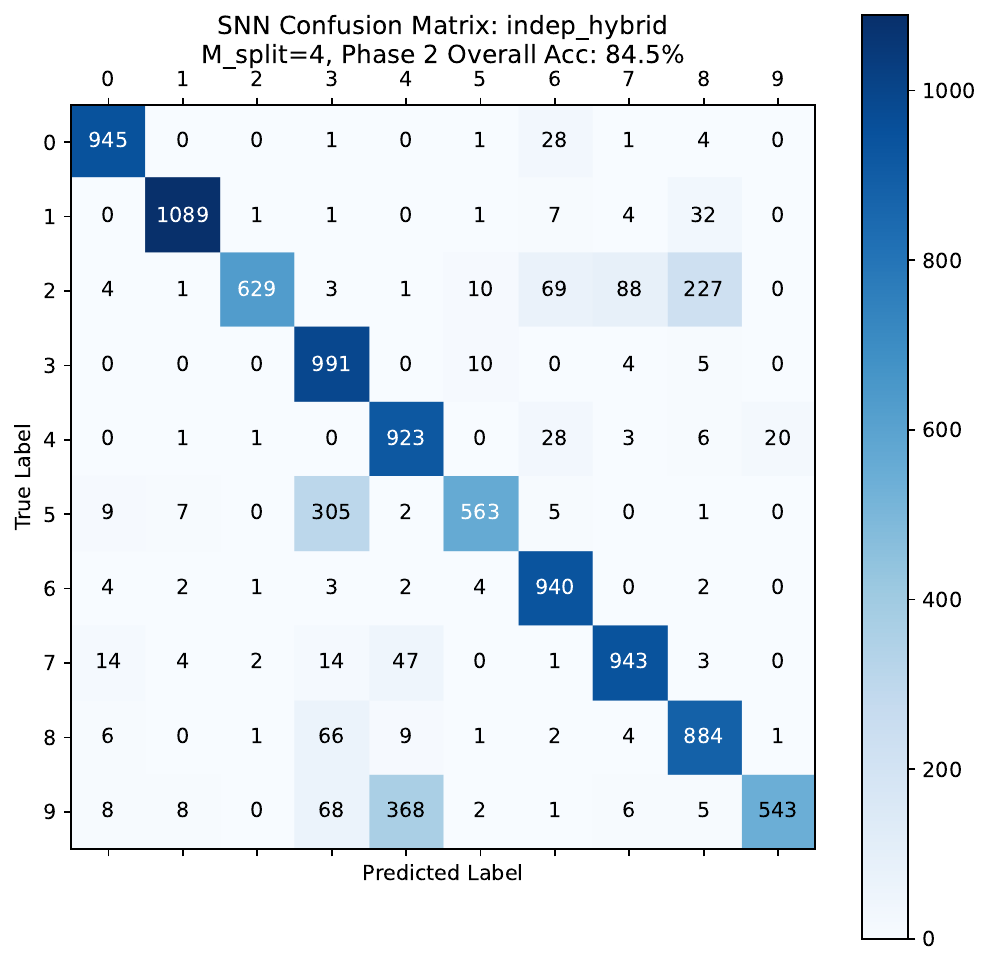}
    \includegraphics[width=0.49\linewidth]{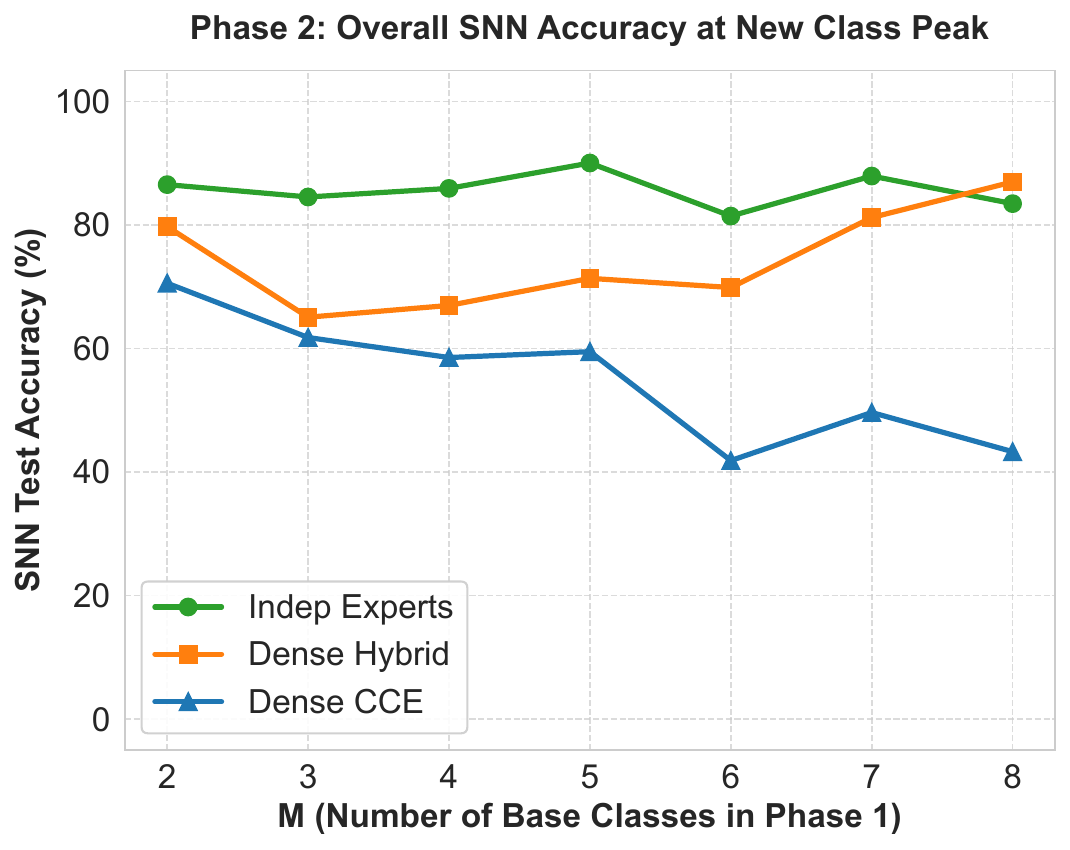}
        \centering
    \includegraphics[width=0.49\linewidth]{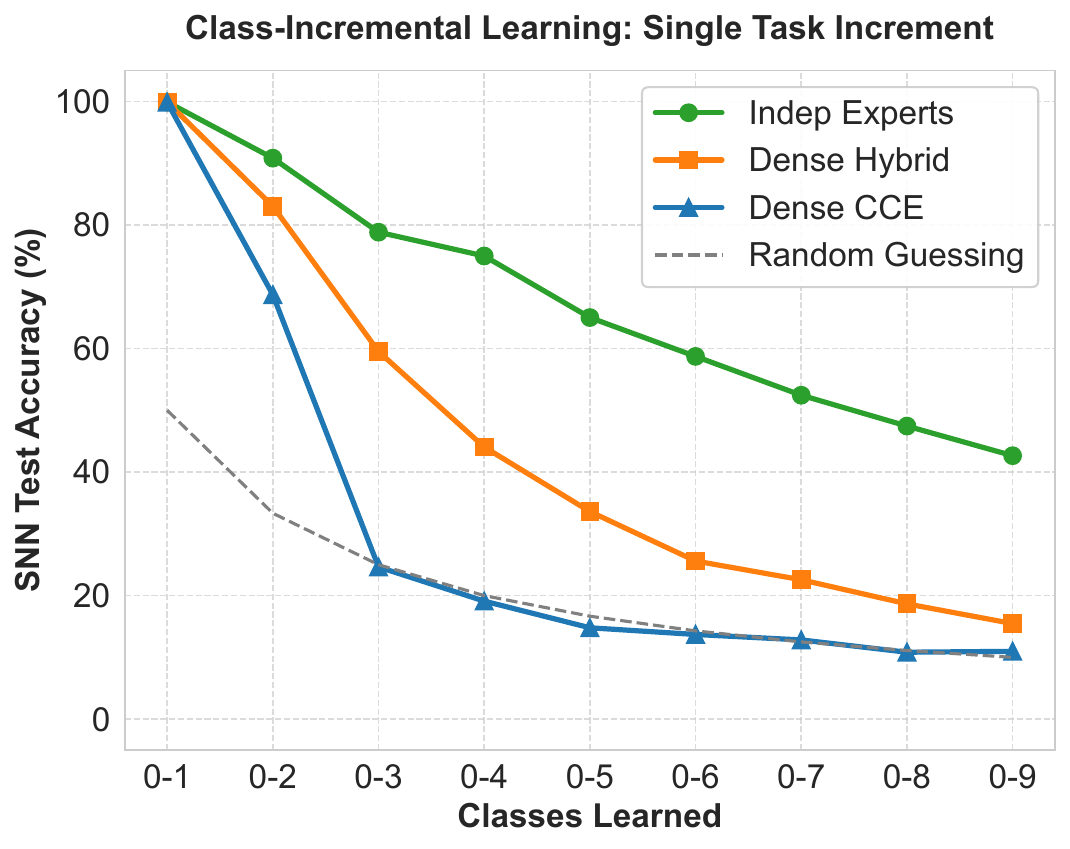}
    \caption{\textbf{Disparity between knowledge retention and overall SNN accuracy.} 
    Top panels: Confusion matrices for continual learning ($M=4$). Dense CCE (left) shows drastic accuracy loss and over-attribution to newly learned classes ($5$--$9$), whereas Independent Experts (right) remain resilient, maintaining clear class separation. 
  Bottom left: Overall SNN accuracy vs. split size $M$. Overall performance in dense models is consistently lower compared with the independent expert architecture, which remains remarkably consistent over all cases. 
 Bottom right: Resilience to catastrophic forgetting during task-incremental learning. By utilizing physically isolated pathways and frozen weights, Independent Experts retain $\sim 40\%$ accuracy at the end of the 10-class sequence. In contrast, dense CCE suffers severe catastrophic forgetting, rapidly collapsing toward random chance ($\sim 1/M$), and dense hybrid's accuracy is in between the two cases. 
 % \\ \textcolor{blue}{[Maybe after submitting, but I would massage the figure to make look nicer. What we can do (I can do it myself if I have the data for the plots): 1) 1 shared colorbar for the top panel; 2) the font size across all sub-figures, 3) Y axes titles can be shared per row of sub-figures]}
    }
    \label{fig:confusion_forgetting}
\end{figure}

The results of our studies are summarized in Figure~\ref{fig:confusion_forgetting}. The top of that Figure illustrates the confusion matrices for the specific case of $M=4$, representing the initial training on classes $0–4$ followed by the incremental learning of classes $5–9$.
The top-left panel displays the results for the Categorical Cross-Entropy (CCE) model, as defined in \eqref{loss_CE}, while the top-right panel shows the performance of the independent experts model. The confusion matrix for the dense hybrid network, trained using the loss function defined by \eqref{loss_def_0}, is omitted for brevity, as its performance closely mirrors that of the independent experts.

A clear distinction in memory retention is observed between the models: while the CCE model's classification is dominated by the most recently learned digits (5–9), the independent experts model exhibits highly robust memory conservation. This model successfully preserves the core representation of Task~1 digits, effectively mitigating the forgetting typically associated with incremental learning.

We further examine the differences in performance across all values of $M$ for the three architectures, as shown in the bottom-left panel of Figure~\ref{fig:confusion_forgetting}. 
The \textit{dense CCE architecture} exhibits catastrophic forgetting of the initial classes, with accuracy dropping to only 20--30\% following the incremental training phase. In contrast, the \textit{independent experts model} maintains high levels of information retention, consistently hovering in the 80--90\% range across all values of $M$. 

The \textit{dense hybrid} case effectively preserves the knowledge of old classes for configurations up to $M=5$ (corresponding to the initial learning of six classes, $0$--$5$). Beyond this point, however, it experiences a significant memory loss. This degradation is attributed to the data imbalance between the two phases, where the loss of interaction between the experts for the new and old classes begins to play a dominant role.

Finally, we tested the Spiking Neural Network (SNN) classification accuracy across single-task incremental learning steps, as presented on the bottom-right panel of Figure~\ref{fig:confusion_forgetting}. In this task-incremental setting, a new class was added at each training step: the network was initially trained on classes 0--1, followed by class 2, class 3, and so on up to class 9. The independent experts architecture demonstrates superior performance compared to both the dense hybrid and dense CCE configurations. By the end of the 10-class procedure, the independent experts model maintains an accuracy of 54.36\%, whereas both alternative methods suffer from catastrophic forgetting, rapidly deteriorating to near-random guessing at approximately 10\%. Notably, the dense hybrid model remained well above the CCE baseline, but fell short of the independent experts. These findings confirm that combining the independent experts architecture with the biomimetic loss function \eqref{loss_def_0} yields exceptional robustness to catastrophic forgetting in continual learning settings
by means of straightforward freezing of the experts' weights from Phase 1.

We further investigated the necessity of synaptic freezing of experts for original digits within the independent experts architecture. For example, for $M=8$,  when experts for previously learned tasks remained plastic during the integration of new classes, accuracy on legacy tasks regressed significantly, to about 40\%, with standard over-attribution to digit '9' similar to Dense CCE in the bottom-left panel in Figure \ref{fig:confusion_forgetting}). This suggests that while structural isolation prevents direct  interference between classes, synaptic stability, mimicking biological memory consolidation, is required to maintain high-fidelity representation across sequential learning stages.  Critically, our D-SNN architecture allows identification of the synaptic weights responsible for previously trained tasks and directly freezing them. This is in contrast to CCE models, where identification of synaptic pathways is difficult and has seen limited success using regularization methods \citep{li2017learning} such as Elastic Weight Consolidation (EWC), Orthogonal Weights Modification (OWM), and Synaptic Intelligence (SI)  \cite{kirkpatrick2017overcoming, zenke2017continual,zeng2019continual}.

%%%% END REM 
\section{Discussion}
\subsection{Biologically Inspired Loss Function Achieves State-of-the-Art Performance}

When developing the D-SNN architecture and "Push-Pull" training, we were inspired by biological neural circuits, where coincidence detection acts as a natural gating mechanism. Only synchronized signals breach the firing threshold to route sensory information and suppress background noise \cite{perez2004intrinsic}. While insect sensory efficiency is traditionally attributed to innate, hardwired architectures \cite{Dielenberg2001, Carde2021} (similar to our independent experts), recent evidence demonstrates that these networks also utilize localized plasticity to develop structured, non-overlapping forward pathways \cite{Tumkaya2022} (similar to our "Push-Pull" training). For instance, training induces inhibitory plasticity within the honeybee antennal lobe (AL) to help it distinguish complex, overlapping odor mixtures \cite{smith2011distributed,lei2022novelty}. By actively suppressing responses to shared chemical compounds while enhancing reward-specific features \cite{joshi2025plasticity}, this remodeling achieves pattern separation through a highly concise neural code.

% Further research on insect nervous systems, such as those of the fruit fly \textit{Drosophila} or the honeybee \textit{Apis}, reveals highly structured neural pathways characterized by striking specificity and limited redundancy. This organization relies heavily on narrow, hard-wired excitatory pathways \cite{winding2023connectome}. Unlike the massive, highly plastic, and densely interconnected general-purpose cortex of mammals, insect brains exhibit a remarkable degree of anatomical and functional separation between circuits responsible for different behaviors. In particular, insect brains exploit two features shaped over millions of years of evolution: (a) \textit{high specificity}, in which neurons are hard-wired to respond selectively to a narrow range of stimuli (e.g., a specific odorant triggering an innate behavioral response, or a defined feature of a courtship display), and (b) \textit{low redundancy}, in which the pathway from sensory input to motor output is physically minimized, favoring speed and extreme energy efficiency over broad generalization.

Research on insect nervous systems (\textit{Drosophila}, \textit{Apis}) reveals highly structured, narrow, and hard-wired excitatory pathways \cite{winding2023connectome}. Unlike the massive, densely interconnected mammalian cortex, insect brains exhibit distinct anatomical and functional separation between behavioral circuits, exploiting two evolutionarily shaped features: 1) \emph{high specificity} (hard-wired neurons selectively respond to narrow stimulus ranges such as specific odorants or courtship displays); 2) \emph{low redundancy} (sensory-to-motor pathways are physically minimized to favor processing speed and energy efficiency over broad generalization). That is, core circuits for innate behavior seem partially predetermined (or genetically encoded) rather than emerging from experience-dependent plasticity. This predetermined architecture is akin to our independent experts in D-SNN.

% Beyond genetically encoded, anatomically fixed pathways, biological brains possess mechanisms to create analogous selectivity dynamically through offline optimization, most notably during sleep. Rapid Eye Movement (REM) sleep is prevalent across species \cite{Peever2017}; however, its precise functional necessity remains debated \cite{Siegel2001}. Recent findings suggest that REM sleep drives memory-specific synaptic weakening \cite{Yang2014,Zhou2020}, accompanied by a shift in the excitatory-inhibitory balance toward inhibition \cite{Tamaki2020}. Together, this suggests that REM sleep reduces interference and sharpens representational separation by suppressing non-essential synapses and promoting the selective routing of neural activity. Crucially, whether selective processing arises anatomically (as in innate insect pathways) or dynamically (through offline synaptic reorganization during sleep), both mechanisms converge on the identical functional outcome: sparse, efficient, and interference-resistant representations.

Beyond predetermined neural architecture, biological brains dynamically generate analogous selectivity through offline optimization during sleep. Although the precise functional necessity of Rapid Eye Movement (REM) sleep remains debated \cite{Siegel2001, Peever2017}, recent findings show that it drives memory-specific synaptic weakening \cite{Yang2014, Zhou2020} and shifts the excitatory-inhibitory balance toward inhibition \cite{Tamaki2020}. This suppresses non-essential synapses, reducing interference and sharpening representational separation. In our case, the proposed "Push-Pull" loss function serves the same purpose as REM sleep. Ultimately, whether selective processing arises anatomically (as in innate insect pathways) or dynamically (via REM sleep), both mechanisms converge on the same functional outcome: sparse, efficient, and interference-resistant representations.

% Recently, several studies have proposed sleep-like unsupervised replay, or Sleep Replay Consolidation (SRC), for artificial neural networks to overcome catastrophic forgetting \cite{tadros2022naturecomm,Kubo2025} and improve generalization \cite{Delanois_ICMLA23,Tadros2020Biologically}. SRC increases sparsity of representations and likely operates similarly to REM sleep, separating synaptic pathways connecting stimulus features to target classification nodes \cite{tadros2022naturecomm,Kubo2025}. While these approaches demonstrate strong performance, they require mapping a trained ANN onto an SNN, executing a computationally expensive offline sleep phase, and mapping back for inference. Inspired by these biological principles, our work demonstrates that path selectivity can be induced directly in artificial networks via architectural constraints and targeted loss functions, bypassing the overhead of a separate sleep phase entirely. 

Recently, Sleep Replay Consolidation (SRC), a sleep-like unsupervised replay, has been proposed to overcome catastrophic forgetting \cite{tadros2022naturecomm, Kubo2025} and improve generalization \cite{Delanois_ICMLA23, Tadros2020Biologically} in artificial neural networks (ANNs). Mimicking REM sleep, SRC increases representational sparsity by separating synaptic pathways connecting stimulus features to target nodes \cite{tadros2022naturecomm, Kubo2025}. However, current SRC approaches require mapping a trained ANN onto a spiking neural network (SNN), executing a computationally expensive offline sleep phase, and mapping back for inference. Inspired by these biological principles, our work demonstrates that path selectivity can be induced directly in artificial networks via architectural constraints by freezing trained experts and allowing new experts to learn new tasks, thereby bypassing the overhead of a separate sleep phase entirely.

Finally, the core of our approach is the Push-Pull loss function \eqref{loss_def_0}, which actively suppresses activity in the pathways of incorrect experts during training, steering the network toward sparse, stimulus-specific routing. This objective is paired with two distinct implementations:  1) the dense hybrid retains full all-to-all connectivity but acquires functional segregation through learned lateral inhibition, allowing modular representations to emerge dynamically; 2) the independent experts model enforces a block-diagonal architecture from the outset, providing structural segregation by construction and eliminating interference entirely. We find that our architecture achieves state-of-the-art performance on standard datasets (MNIST, Fashion-MNIST, CIFAR) and shows strong resilience to catastrophic forgetting in class-incremental learning. Furthermore, the independent experts model achieves this with only a fraction of the parameters used by competitive models. 

The three-way comparison between CCE training, the dense hybrid, and independent experts allows the contributions of the loss function and architectural constraints to be disentangled. Since the dense hybrid receives the same Push-Pull objective $\mathcal{L}_{PP}$ as the independent experts but retains full all-to-all connectivity, the performance difference between CCE and the dense hybrid reflects the loss function alone, while the further gain of independent experts over the dense hybrid reflects structural segregation. The results confirm that both factors contribute independently: the Push-Pull loss alone induces functional modularity in a fully connected network, while structural isolation compounds these gains by eliminating residual crosstalk that learned inhibition cannot fully suppress.

The dense hybrid results also quantify the metabolic cost of dynamic versus structural isolation. To approximate the block-diagonal structure that independent experts enforce by design, the dense hybrid develops strong lateral inhibitory weights, consuming parameters solely to cancel activity that never arises in the independent experts case (Figure \ref{fig:weights_dense_OvA_PP}). This difference in number of synaptic operations and firing rate is 
demonstrated in the performance and efficiency results presented on Table \ref{tab:combined_sparsity_measurements}). In settings where the class structure is not known prior to training and a block-diagonal architecture cannot be pre-specified, the dense hybrid model provides a flexible path to functional segregation at the cost of metabolic efficiency.

\subsection{Advantages of Class-Specific Path-Based Architectures}

Dedicating structurally isolated neuronal groups to specific input classes offers profound evolutionary and computational advantages, translating directly to the design of next-generation neuromorphic hardware:

\begin{enumerate}
    \item  \textbf{Resource and Energy Efficiency}: The biological brain avoids allocating vast synaptic resources to general-purpose computation; instead, limited resources are dedicated to high-priority survival tasks. This mirrors the design philosophy of specialized neuromorphic accelerators for edge AI, where strict metabolic and parameter efficiency are necessary. Due to the inherent sparsity of the D-SNN framework, our method is substantially more energy efficient than dense networks of equivalent accuracy. In particular, training becomes substantially faster. Also, by keeping non-target experts metabolically quiescent, our approach creates opportunities to construct networks with extremely high energy efficiency, which is vital for edge computing applications.

    \item \textbf{ Mitigation of Catastrophic Forgetting:} Biological sensory-motor systems exhibit high degrees of functional compartmentalization \cite{heinze2011anatomical, stone2017anatomically}. This physical separation prevents cross-talk between competing sensory modalities. Mirroring this architecture, our independent experts framework ensures that the optimization of weights for a novel class (e.g., digit `9') is mathematically decoupled from the synaptic weights of prior classes. By enforcing this structural modularity, we mitigate the weight-interference patterns that typically drive catastrophic forgetting in homogeneous architectures.

    \item \textbf{Traceability and Verifiability:} In fully entangled ANNs, the highly entangled nature of representations creates a "black box" that hinders certification for safety-critical applications. By contrast, the D-SNN offers strict traceability of signal. Because class-specific logic is localized to physically isolated pathways, activation patterns can be deterministically audited. If the network triggers a specific classification, the exact sub-circuit responsible is explicitly known, making the system highly transparent and verifiable.  This is crucial for quality control and troubleshooting of prediction artifacts (e.g., systematic errors or bias in class predictions). That is, the "broken" expert can be easily located, retrained, and calibrated more quickly than the entire network.
    
\end{enumerate}

\subsection{Considerations regarding Modularity and Scalability}
By physically severing lateral connectivity between decision pathways, we mathematically constrain the system to operate as a predictable ensemble of discrete functions. This ensures that as the network scales, the experts remain functionally isolated and their actions fully explainable. This property becomes critical when the number of classes increases significantly, such as in the classification of Chinese characters in the CASIA dataset, which exceeds 7,000 distinct categories \cite{liu2011casia}. 

To scale the D-SNN for such tasks, one approach is to increase the internal capacity of each expert while maintaining their independence. Alternatively, one could allow sparse, dynamically routed connectivity between experts, similar to a \textbf{Mixture of Experts (MoE)} architecture \cite{jacobs1991adaptive, shazeer2017outrageously}. Even in this configuration, the architecture remains strictly feed-forward and its decision pathways remain highly traceable, avoiding the opaque ``black box'' entanglement typical of dense models. Such an architecture is exceptionally well-suited for high-stakes applications where AI decisions bear significant responsibility and must remain fully auditable regardless of system complexity. Furthermore, the inherent energy efficiency and sparsity of our methodology make it particularly well-suited for mission-critical edge applications where computational resources are constrained but reliability and explainability are paramount.

%\textcolor{blue}{[TBD] Further success of our architecture presented in this paper aligns with the Lottery Ticket hypothesis \cite{frankle2018the}, providing an explicit way to radically lower the number of parameters in a network without sacrificing the performance. }

\subsection{Conclusions}
Inspired by the structured, non-overlapping neural pathways observed in biological sensory systems, we introduced the Decomposable Spiking Neural Network (D-SNN) framework, which combines a block-diagonal independent experts architecture with the Push-Pull loss function to enforce sparse, stimulus-specific routing directly through training. This biologically motivated design yields a principled solution to two long-standing challenges in neuromorphic computing: catastrophic forgetting in continual learning and metabolic inefficiency in dense networks. Our results demonstrate that structural segregation, rather than costly learned inhibition, is the more effective and efficient strategy for modular representation, achieving state-of-the-art accuracy on standard benchmarks with a fraction of the parameters of competing models. Together, these findings suggest that explicitly encoding biological organizational principles into artificial network architectures offers a promising path toward scalable, energy-efficient, and continually adaptable neural systems.

\section{Methods}

To contextualize the architectural and metabolic advantages of the Decomposable SNN (D-SNN), we first outline the fundamental mechanics of Spiking Neural Networks and the theoretical difficulties inherent to their optimization. For a comprehensive review of modern SNN training methodologies, see \cite{eshraghian2023training}.

\subsection{The Leaky Integrate-and-Fire (LIF) Model}

The foundational computational unit in a Spiking Neural Network (SNN) is the  \LIF{} model. This model approximates the complex biological membrane dynamics by a streamlined first-order differential equation describing the evolution of the membrane potential $V_m$ over time $t$:
\begin{equation}
\tau \frac{dV_m}{dt} = -(V_m - V_{\text{rest}}) + R I(t) \, , \quad V_m \rightarrow V_{\text{rest}} \mbox{ if } V_m > V_t \, . 
\label{LIF}
\end{equation}
where $\tau$ is the membrane time constant, $V_{\text{rest}}$ is the resting potential, $R$ is the membrane resistance, and $I(t)$ represents the integrated synaptic input current. If $V_m$ exceeds the firing threshold $V_t$,  the membrane potential $V_m$ resets to the rest potential $V_{\text{rest}}$ (or, in some models, a potential higher than the resting potential).  

In our implementations, we discretize the  continuous dynamics given by \eqref{LIF} using discrete  time steps $\Delta t$, resulting in an iterative update rule for the membrane potential:
\begin{equation}
V[t+1] = \beta V[t] + \kappa \sum_j W_j  S_j(t) - V_{\text{th}} S_{\text{out}}\, , \quad 
 S_{\text{out}} = 
\begin{cases}
0 & \text{if } V[t] < V_{\text{th}} \\
1 & \text{if } V[t] \geq V_{\text{th}} \, . 
\end{cases}
\label{LIF_updateRule}
\end{equation}
where $\beta = e^{-\Delta t / \tau}$ is the \emph{membrane decay constant} (representing the persistence of the potential), and $\kappa$ is the \emph{synaptic efficacy}, a gain factor that scales the impact of incoming spikes $S_j(t)$ weighted by their weights $W_j$. When $V[t]$ exceeds a defined threshold $V_{\text{th}}$, the neuron emits a discrete spike, and the potential is reset, illustrated by the function $\mathcal{R}(t)$.

The steady-state response of this discrete system is characterized by the \textbf{effective resistance} $R_{\text{eff}}$, which governs the total integration gain:
\begin{equation}
R_{\text{eff}} = \frac{\kappa}{1 - \beta}
\label{LIF_Reff}
\end{equation}
In our implementation, we set $\kappa = 0.7$, ensuring that the $R_{\text{eff}}$ maintains the network within a sensitive dynamic range while preventing hyper-excitability. The value of $V_{\text{th}}$ is always taken to be 1 in all implementations. We typically take $\beta = 0.6$, giving $\tau=\Delta t/\log(\beta^{-1}) \simeq 1.96 \Delta t$.
Despite its mathematical simplicity, the \LIF{} model captures the essential biological properties of temporal integration and threshold-based spiking.

\subsection{Information Encoding and Decoding in Spiking Networks}
Unlike traditional \DNN{}s, which propagate continuous-valued floating-point activations, SNNs inherently process information via asynchronous, discrete temporal events (spikes). Consequently, input data is typically transformed into spike trains using either rate coding (where information is represented by spike frequency) or temporal coding (where information is embedded in precise spike timing) \cite{eshraghian2023training}. However, because our methodology relies on training a surrogate continuous-valued ANN and subsequently converting it to an SNN, we bypass these traditional stochastic encoders. Instead, we employ direct input encoding: the static input (e.g., an image of a digit) is treated as a stationary signal applied continuously across all time steps of the SNN simulation. This approach perfectly mirrors the static analog inputs used during the ANN training phase, ensuring a faithful 1:1 mathematical mapping of the network's behavior and allowing us to properly evaluate the deterministic accuracy of the D-SNN.

%%%% END REM 
\subsection{The Lack of Differentiability in SNN Learning}

The inherent discontinuity of the \LIF{} spiking mechanism represents a challenge to the application of gradient-based optimization. The emission of a spike function $S(t)$ is governed by the Heaviside step function:
$$
S(t) = \begin{cases}
1 & \text{if } V_m(t) \ge V_{\text{th}} \\
0 & \text{otherwise}
\end{cases}
$$
The derivative of this function is zero everywhere except at $V_{\text{th}}$, where it is undefined (a Dirac delta). This "dead gradient" physically prevents the application of Backpropagation Through Time  (BPTT), the standard optimization engine for recurrent systems, which strictly requires continuous, non-zero gradients to assign credit across hidden layers.

Modern techniques to circumvent this limitation generally fall into two categories:
\begin{itemize}
    \item \textbf{Surrogate Gradients (SG):} During the backward pass of BPTT, the true, undefined gradient of the step function is replaced by a continuous, piece-wise differentiable approximation (the ``surrogate''). This mathematical workaround permits the calculation of gradient descent updates and currently yields state-of-the-art accuracy in deep SNNs. However, SGs introduce complex hyperparameters and often provide an imperfect mapping of the true error landscape.
    
    \item \textbf{ANN-to-SNN Conversion:} A standard \DNN{} (typically using ReLU activations) is trained conventionally, after which its weights are mapped to an SNN by equating the continuous ReLU output to the discrete firing rate of a \LIF{} neuron. While this yields high accuracy, it sacrifices the intrinsic temporal dynamics and event-driven efficiency of native SNNs. In our work, we employ a carefully designed version of this approach that takes advantage of its simplicity and efficiency, while capturing the essential dynamics due to the architecture design and the biomimetic loss function \eqref{loss_def}.
\end{itemize}

\subsection{Metabolic and Architectural Constraints of Standard Optimization Procedures}

Beyond the challenge of non-differentiability, wide-scale application of neuromorphic computing, especially in edge autonomy, needs methods of learning that are substantially more computationally efficient than BPTT and similar methods. The current methods for learning in SNNs suffer from the following drawbacks.
\begin{enumerate}
    \item \textbf{Prohibitive Memory Footprint:} To compute gradients in the backward pass, methods like BPTT requires remembering the membrane potentials and spike events of every neuron at every timestep during the forward pass. This makes the training procedure computationally expensive (requiring powerful GPUs), and hard to implement on actual neuromorphic chips.
    
    \item \textbf{Catastrophic Forgetting in Online Learning:} When a neuromorphic device, which was trained offline, tries to adapt to the changes in the environment, it needs to augment the weights. However, the global nature of weight updates corrupts previously tuned synapses, triggering catastrophic forgetting \cite{french1999catastrophic, hayes2021replay, luo2025empirical}.
\end{enumerate}
Therefore, the realization of functional, autonomous edge neuromorphic AI requires a novel approach. Rather than forcing globally entangled, memory-intensive algorithms onto constrained hardware, we will design structurally decomposed topologies, that inherently support localized, interference-free adaptation.

\subsection{Surrogate Activation and Network Architecture}

To bridge the gap between continuous gradient-based optimization and discrete spiking inference, we model the neuron activation during the ANN training phase using a continuous, differentiable surrogate function. Specifically, we approximate the discrete firing rate based on the voltage surplus $\Delta V = V - V_{\text{th}}$ using a steep sigmoid function:
\begin{equation}
    \Phi(V) = \sigma\left(\frac{\Delta V}{\epsilon}\right) = \frac{1}{1 + e^{-\Delta V / \epsilon}}
    \label{sharp_sigmoid}
\end{equation}
where $\epsilon$ controls the steepness of the surrogate gradient. 

Because our experimental focus is on high-dimensional image classification, the base architecture of our networks, which is identical for both the surrogate ANN and the true SNN,  is structured as follows:
\begin{enumerate}
    \item \textbf{Input Layer:} Encodes the raw pixel intensities.
    \item \textbf{Shared Receptor:} Comprises one (for MNIST/FMNIST) or two (for CIFAR-10/100) trainable $3\times3$ convolutional layers.
    \item \textbf{Hidden Layers:} We use one of the following three architectures: 
    \begin{enumerate}
    \item \textbf{Dense Architecture:} 
    \emph{Dense Hybrid} or \emph{Dense CCE}: Two dense layers of dimensions $K \cdot N_1$ and $K \cdot N_2$, respectively, where $N_1$ and $N_2$ represent the number of neurons allocated per class, and $K$ is the total number of distinct classes (\emph{e.g.}, $K=10$ for MNIST). The architecture for these two cases is identical; the difference is in learning procedure. For Dense Hybrid, we use the biomimetic loss function \eqref{loss_def} for learning, whereas Dense CCE case uses the standard Categorical Cross-Entropy \eqref{loss_CE}, as explained in Section~\ref{sec:learning_procedure} below.
        \item \textbf{Independent Experts (or Indep Experts):} In this case, the hidden layers consist of $K$ isolated experts without cross-connections between each other, and each one has two hidden layers of dimensions $N_1$ and $N_2$. An example of such an architecture is shown in Figure~\ref{fig:experts}.
        \end{enumerate}
    \item \textbf{Output Layer:} $K$ neurons, densely connected to the final hidden layer, mapping the latent representations to class predictions.  Note that for the independent experts, each output neuron is connected to its isolated branch only.
\end{enumerate}

% We show that training a surrogate ANN using the biomimetic loss function \eqref{loss_def} below, in the case of the independent expert architecture, yields a network that is accurate, efficient, and robust. 

\subsection{The Biomimetic Loss Function for Pathway Isolation}
\label{sec:learning_procedure}
To induce the emergence of class-specific processing pathways, we conceptually partition the hidden dense layers into $K$ distinct sub-networks, or `experts' $\mathcal{B}_k$. For a given class $k$, its corresponding expert $\mathcal{B}_k$ consists of $N_1$ and $N_2$ neurons in the first and second hidden layers, respectively. For instance, the expert designated for class $k=4$ (where $K=10$) encompasses all neurons with indices $3N_1 < \alpha \leq 4N_1$ in the first layer.

Suppose we have $M$ training examples, and a specific sample $m$ belongs to the ground-truth class $k_m$. For each forward pass, we compute the total post-activation neural output within the target (true) expert and the non-target (false) experts. Let $x_\alpha$ denote the total integrated synaptic input to neuron $\alpha$, and $z_\alpha = \Phi(x_\alpha)$ denote its continuous post-activation output. We define the true and false activity scores as:
\begin{equation}
    S_{\rm true}^m = \sum_{\alpha \in \mathcal{B}_{k_m}} z_{\alpha} \, , \quad 
    S_{\rm false}^m = \sum_{\alpha \notin \mathcal{B}_{k_m}} z_{\alpha}. 
    \label{score_def}
\end{equation}
Since the surrogate activation $\Phi(V)$ is strictly positive, the scores $S_{\rm true}^m$ and $S_{\rm false}^m$ are bounded above zero. To enforce spatial routing, we formulate the novel `Push-Pull' loss function:
\begin{equation}
    \mathcal{L}_{\text{PP}} =\sum_{m=1}^M \log \left(S_{\rm false}^m+ S_{\rm true}^m \right) - \log\left(S_{\rm true}^m \right),
    \label{loss_def}
\end{equation}
which is the mathematically precise version of equation \eqref{loss_def_0}. Crucially, this formulation introduces no additional tunable hyperparameters. $\mathcal{L}_{\text{PP}}$ directly optimizes the spatial distribution of internal metabolic activity rather than the final output logits. It explicitly minimizes neural activity in all non-target pathways ($S_{\rm false}^m$) while maximizing it in the target pathway ($S_{\rm true}^m$). This active suppression explicitly carves out sparse, non-overlapping routing structures for neuron signal, an effect that is impossible to achieve in standard optimization procedures, which we use as a baseline. 

For the dense hybrid model (non-isolated experts) that we show as an alternative and train using our Push-Pull loss \eqref{loss_def}, we manually choose non-overlapping 'correct' sub-networks of neurons to calculate the term $S_{true}^m$ for every class $k$ (similar but not identical to Figure~\ref{fig:experts}). However, in this case, we retain lateral synaptic connections such that the sub-networks are not isolated, i.e. the last dense layers in Figure~\ref{fig:experts} become entangled between all classes. In other words, the Push-Pull loss \eqref{loss_def} imposes a soft contraint to suppress the crosstalk between the correct and wrong sub-networks. 

\subsection{Comparing to the Categorical Cross-Entropy Baseline}

To benchmark the efficacy of the Push-Pull mechanism, we contrast it against the standard Categorical Cross-Entropy (CCE) loss. Consider a dataset consisting of inputs $\mathbf{x}^{n}$ and the ground truth class label outputs $c^{n}$, with $n=1, \ldots, N$ comprising $N$ independent observations. Here, $c^{n}$ is simply an integer indicating the class, so $c^{(n)} \in \{1, \dots, K\}$ . For each input, the network generates a vector of raw output logits $\mathbf{z}(\mathbf{x}^{(n)}) \in \mathbb{R}^K$ (a vector of real numbers), which is mapped to a normalized probability distribution $\mathbf{p}^{n}$ via the Softmax function:
\begin{equation}
    p_k^{n} = \frac{\exp(z_k^{n})}{\sum_{l=1}^{K} \exp(z_l^{n})} \, , \quad k = 1, \dots, K \, .
    \label{softmax}
\end{equation}
The total CCE loss over the dataset is defined as the empirical mean of the negative log-likelihood. For each data sample, $n$, we define a $K$-dimensional vector $\mathbf{y}^n$ (a target vector) which has $1$ at the component $c^n$ and $0$ everywhere else. Then, the CCE loss is defined as: 
\begin{equation}
    \mathcal{L}_{\text{CCE}} =
    - \frac{1}{N} \sum_{n=1}^N \sum_{k=1}^K y_k^n \log p_k^{n}   = - \frac{1}{N} \sum_{n=1}^{N} \log(p_{c^{n}}^{n})
    \label{loss_CE}
\end{equation}
Minimizing $\mathcal{L}_{\text{CCE}}$ is mathematically equivalent to performing Maximum Likelihood Estimation (MLE) for the network weights.  One may notice that equations \eqref{loss_def} and \eqref{loss_CE} are structurally similar, with one important difference. Specifically, while CCE \eqref{loss_CE} measures the difference of logarithms of output logits between correct and wrong \emph{labels}, $\mathcal{L}_{\text{PP}}$ \eqref{loss_def} measures the difference of logarithms of total neural activity between correct and wrong \emph{experts}.

While CCE \eqref{loss_CE} effectively aligns the final outputs with the target labels, it does not enforce anything related to the internal architecture; it optimizes solely for final output behavior without imposing constraints on signal routing within hidden layers. This lack of structural regularization encourages dense structural entanglement of neural signal, where hidden neurons indiscriminately support multiple class outputs to satisfy the global loss minimization. This leads to diminished traceability and the phenomenon of catastrophic forgetting. In contrast, our Push-Pull mechanism \eqref{loss_def} and the neural network architecture enforces a functional 'division of labor', ensuring that experts remain independent and functionally isolated across during learning and inference.

\paragraph{Summary of Architectural Configurations}

Throughout our experiments, we consistently evaluate three distinct architectural regimes:
\begin{enumerate}
    \item \textbf{Dense CCE:} A standard, fully connected architecture optimized using the global CCE loss function \eqref{loss_CE}.
    \item \textbf{Dense Hybrid:} A fully connected architecture optimized using the Push-Pull loss function \eqref{loss_def}, which actively penalizes crosstalk but retains lateral synaptic connections.
    \item \textbf{Independent Experts:} Our proposed decomposable topology, where lateral connections between the $K$ experts are physically severed (structural isolation), and the network is optimized using the Push-Pull loss function \eqref{loss_def}.
\end{enumerate}

\subsection{ANN-to-SNN Porting and Inference Validation}

Following the optimization of weights in the continuous ANN domain, we validate the architecture's true performance by directly mapping the learned parameters onto a discrete SNN. The SNN utilizes \LIF{} neurons  described by  \eqref{LIF} - \eqref{LIF_Reff}, with a fixed firing threshold $V_{\text{th}} = 1.0$, a membrane potential decay factor $\beta = 0.5$, and an effective resistance $R_{\rm eff}=0.7/(1-\beta)$. 

During validation, the SNN is simulated over $T=100$ discrete timesteps. Classification is determined dynamically by identifying the output neuron that emits the maximum number of spikes. Thus, while learning procedure utilizes a highly efficient surrogate ANN, all reported accuracies are derived from SNN simulations only. This hybrid learning/porting procedure is highly computationally efficient. A complete training and porting cycle requires only several hundred epochs, completing within minutes on standard consumer hardware (e.g., an Apple M2 Max architecture). In contrast, training an equivalent topology exclusively in the spiking domain via Backpropagation Through Time (BPTT) requires orders of magnitude more computational overhead, typically necessitating dedicated high-performance GPUs in HPC computing clusters.

\subsection{Network Architectures for MNIST and Fashion-MNIST}

To evaluate the proposed loss function and structural decomposability, we designed specific network topologies for the MNIST and Fashion-MNIST (FMNIST) datasets. In all experiments, the network is fundamentally separated into two parts: it begins with a convolutional receptor block, which we call retina, to extract spatial features, followed by a partitioned ensemble of parallel expert pathways in the case of independent experts, or a dense network in the case of CCE or dense hybrid. 

\paragraph{MNIST Architecture}
For the MNIST dataset of handwritten digits, the retina consists of a single trainable convolutional layer comprising eight $3 \times 3$ filters ($F=8$), followed by average pooling. For the case of independent experts, the latent representation from retina is then routed to an ensemble of $K=10$ structurally isolated expert networks, each consisting of two hidden layers $N_1=64$ and $N_2=32$. 

Because the independent experts lack lateral synaptic connections, the weight matrices governing the hidden layers are strictly block-diagonal. This complete independent experts configuration contains a total of 271,768 trainable parameters. By contrast, a baseline dense architecture, permitting all-to-all connectivity across the equivalent pool of hidden neurons, requires 458,968 parameters. The structural isolation thus reduces the required synaptic resources by a factor of two.

\paragraph{Fashion-MNIST (FMNIST) Architecture}
The FMNIST dataset presents a greater challenge than MNIST, requiring the network to resolve more complex morphological shapes and textures. To accommodate this, we scaled the network capacity while strictly maintaining the Independent Expert design. 

The capacity of the retina filters was expanded to $F=16$  trainable convolutional filters, with extended scaling experiments, not shown here, testing up to $F=32$ filters. The hidden dense layers were partitioned in such a way that each of the 10 expert pathways contains $N_1=64$ neurons in the first hidden layer and $N_2=32$ neurons in the second hidden layer. Consistent with the MNIST setup, this structurally isolated network requires approximately 522,000 parameters, preserving the 10-fold reduction in parameter footprint compared to the fully dense counterpart while providing sufficient representational capacity for the more complex FMNIST feature space.

\subsection{CIFAR-10 and CIFAR-100 Experimental Configuration}

\paragraph{Dataset Preprocessing}
The CIFAR-10 and CIFAR-100 datasets, comprising $32 \times 32$ color images, were normalized using a mean and standard deviation of 0.5 for each RGB channel. This zero-centering of the input distribution is critical for preventing early membrane potential saturation in the spiking neurons and ensuring balanced gradient propagation during the initial stages of optimization.

\paragraph{CIFAR-10 Architecture and Scaling}

To accommodate the increased spatial complexity of natural image data, we utilized a tiered architecture consisting of a shared feature extractor followed by the isolated expert modules. 

\begin{itemize}
    \item \textit{Convolutional Receptor Block (Retina):} The shared sensory front-end is expanded into a deeper three-layer convolutional topology adapted from Cao et al. (2015). Layers 1 and 2 utilize $5 \times 5$ kernels with a stride of 1 and padding of 2, each structurally coupled to a $2 \times 2$ max-pooling layer with a stride of 2. Layer 3 utilizes a $3 \times 3$ kernel with a stride of 1 and padding of 1. The filter density across all three layers is held uniform and governed by a base scaling factor $F = 64$. This progression systematically compresses the spatial dimensions of the input down to an $8 \times 8$ feature map. The resulting flattened feature vector of size $64F$ is fed to the hidden layers.
    
    \item \textit{Hidden Layer Configurations:} Following the convolutional block, the partitioned latent features are routed into one of three comparative topologies:
    \begin{itemize}
        \item \textit{Independent Experts:} The architecture features ten isolated structural experts, where each individual expert is exclusively dedicated to processing a single class. Every expert consists of two localized hidden layers with highly constrained, fixed neuron capacities of $n_1 = 5$ and $n_2 = 5$. 
        \item \textit{Dense CCE (Baseline):} A standard, fully connected architecture configured to match the total aggregate capacity of the expert ensemble ($5M = 50$ neurons in the first hidden layer and $5M = 50$ in the second). It is optimized via Categorical Cross-Entropy \eqref{loss_CE}. 
        \item \textit{Dense Hybrid:} Utilizes the exact same fully connected topology and aggregate neuron counts ($50$ and $50$) as the Dense CCE baseline, but is optimized using the joint Push-Pull loss function \eqref{loss_def}.
    \end{itemize}

    \item \textit{Spiking Neuron Model:} For discrete SNN verification, we utilized a \LIF{} model with a membrane potential decay factor $\beta = 0.5$ and a uniform firing threshold $V_{\text{th}} = 1.0$. To maintain deterministic behavior across an extended $T=100$ timestep inference window, neurons execute a strict hard reset-to-zero mechanism immediately following spike emission. Continuous weight optimization within the surrogate ANN framework is driven by a Sharp Sigmoid activation function defined with a smoothing parameter of $\epsilon_{\text{sigmoid}} = 0.2$.
\end{itemize}
We used data augmentations (translations and mirror images) to enhance the learning procedure. The total capacity for each model is approximately 352K parameters for the fully dense configurations and 349K active parameters for the independent experts model, of which roughly 144K is allocated to the convolutional front-end. Notably, the difference in parameter footprint between the fully dense and independent expert models increases quadratically as the internal size of the individual expert hidden layers scales up.

Models were optimized for 200 epochs using the Adam optimizer with a training batch size of 200. To prevent instability and guarantee high-confidence convergence, the learning rate followed a step decay schedule, starting at 0.001 and decreasing by a factor of 2 every 50 epochs. Network scaling and sparsity behaviors were characterized around a baseline filter tier of $F = 64$. Weight optimization was performed exclusively via a continuous surrogate ANN using the designated objective functions.

% This allowed for complete training and SNN verification within approximately 10 minutes on standard consumer hardware (Apple M2 Max laptop), bypassing the need for dedicated high-performance computing clusters.

\paragraph{CIFAR-100 Architecture and Class Scaling}
To test the limits of structural decomposability under extreme class-scaling conditions, we adapted the architecture for subsets of the CIFAR-100 dataset. We evaluated both the dense hybrid architecture (a fully connected matrix trained via the Push-Pull loss \eqref{loss_def} to induce functional sparsity) and the strictly isolated independent experts topology.

The retina (convolutional preprocessing layer) was expanded to a three-layer convolutional block with $F=64$ filters. This block feeds into two hidden dense layers with a fixed global width of 500 neurons. To systematically evaluate performance as a function of task complexity, simulations were conducted for an increasing number of target categories $M$, where $M \in \{10, 20, \dots, 100\}$. For the independent experts configuration, the 500 hidden neurons were evenly partitioned among the $M$ classes, such that the number of neurons per expert layer was strictly defined as $N_1 = N_2 = \lfloor 500 / M \rfloor$.

\subsection{Continuous Learning Experimental Protocol}
To evaluate the robustness of the proposed architecture against catastrophic forgetting during incremental learning, we designed a sequential class-addition protocol using the MNIST dataset. The experiment compares the retention capabilities of the structurally isolated independent experts against the fully entangled Dense CCE and dense hybrid baselines. 
The sequential learning task is divided into two phases:

\paragraph{Phase 1: Base Optimization} 
The networks are initially trained exclusively on a subset of the data comprising digits 0 through $M$ ($M+1$ class), for $M=2,\ldots,8$.  All the architectures utilize a single-layer convolutional shared receptor with $F=8$ filters, and two hidden layers configured to an aggregate capacity of $10 \cdot N_1$ and $10 \cdot N_2$ neurons, where $N_1 = 64$ and $N_2 = 32$. During this phase, all network architectures undergo standard optimization until convergence is achieved, with the accuracy close to 100\% for all cases.  

\paragraph{Phase 2: Incremental Class Addition}
Following the initial optimization learning digits $0-M$, the networks are tasked with learning new digits $M+1,\ldots,9$, without having access to the full  dataset for the old digits 0-8. To simulate a constrained memory buffer, we generate a synthetic rehearsal dataset for the base classes. For each digit $0-M$, we compute the  average of all Phase 1 training images, resulting in $M+1$ images. We then generate synthetic samples by injecting $15\%$ random noise into these $M$ average images. The number of synthetic images generated per base class from each average image for each class is set to be equal to the number of true training images available for the new digit `9`, around 6,000 for the MNIST dataset. 

\paragraph{Phase 2 Training}
The networks resume training using only the true images for the new classes and  synthetic noisy average images of the old classes. Crucially, the architectural models are subject to different optimization constraints during this phase:
\begin{itemize}
    \item \textbf{Independent Experts and Dense Hybrid:} All synaptic weights within the expert modules for digits $0-M$ are frozen. Gradient updates are applied only to the newly activated expert pathway designated for digits $M+1$ to $9$.
    \item \textbf{Dense Architectures (CCE):} Because dense architectures inherently rely on distributed, globally interacting networks, one cannot simply isolate weight freezing by class. Therefore, all synaptic weights across the entire network remain unfrozen and are subject to continuous gradient updates during Phase 2. This factor makes the dense CCE case much more susceptible to catastrophic forgetting. 
\end{itemize}

\paragraph{Task-Incremental Continual Learning Protocol}
To further investigate the capabilities of our architecture in an extended continual learning scenario, we evaluate the model using a step-by-step task-incremental learning setting \cite{de2021continual}. In this experiment, the training procedure begins by learning a base task consisting of digits 0 and 1. We then incrementally introduce subsequent digit classes one by one, up to digit 9. 

During the introduction of each new task, the model is trained on the authentic images of the new digit alongside a highly compressed pseudo-rehearsal dataset. This replay dataset consists of synthetic, noise-injected images generated from a single averaged representation of each previously learned digit. To accommodate the new information sequentially without structural bottlenecking, the network dynamically expands at each incremental step, adding $N_1=64$ neurons to the first hidden layer and $N_2= 32$ neurons to the second. Consistent with the independent and dense experts paradigm, the weights in the previously trained, 'old' parts of the hidden layers are explicitly frozen at each step to protect the learned representations, whereas the weights in dense CCE case remain globally unfrozen.

\paragraph{Hyperparameters and Computational Setup}
The experimental framework is governed by a consistent set of hyperparameters across all learning phases to ensure reproducibility. For the initial base training (digits $0-M$ in the first case or digits $0-1$ in the second case of incremental learning), models were trained for $100$ epochs using the Adam optimizer with an initial learning rate of $5 \times 10^{-4}$ for the CCE baseline and $5 \times 10^{-4}$ for hybrid configurations, with the learning rate halved every $50$ epochs. During subsequent incremental steps (digits $2-9$), each new class was trained for $100$ epochs at a constant learning rate of $10^{-3}$. The mechanism creating "memory" images from the averaged digits in the previous learning procedure utilized a noise fraction of $0.15$ applied to the mean images. All simulations utilized a batch size of $512$. 

The SNN conversion parameters were held constant to ensure valid verification: the membrane decay ($\beta$) was set to $0.6$, the firing threshold ($V_{thresh}$) to $1.0$, and the sigmoid steepness ($\epsilon$) to $0.2$ (see eq. \ref{sharp_sigmoid}). For temporal dynamics, the SNNs were simulated over $T=100$ time steps per inference, with an effective resistance ($R_{eff}$) calibrated to $0.7 / (1.0 - \beta)$ to ensure robust spike propagation.

Consistent with all prior protocols, performance evaluation and final classification accuracy are  evaluated via direct SNN simulation rather than the continuous ANN surrogate.

\subsection*{Declaration on the use of AI in manuscript preparation}
During the preparation of this work, the authors used \emph{Gemini 3 Pro} to streamline manually written source code for readability and computational efficiency. The authors also used this tool to refine the stylistic flow and readability of the manuscript. Following the use of this service, the authors thoroughly reviewed, verified, and edited the content as needed, and the authors take full responsibility for the integrity of the final code and manuscript text.

\section*{Data and Code Availability}
We have used standard data sets MNIST, Fashion-MNIST and CIFAR-10/100. These data sets are publicly available. 

The final version of the code used in this paper is available from the following link: 
\url{https://github.com/vputkaradze/Neuromorphic_Independent_Experts}

\section*{Author Contributions}
All authors contributed to the conceptualization and framing of the study. V.P. and S.G. developed the implementation of the architecture, produced the code and figures, performed the formal analysis and data validation. V.P., S.G., and M.B. drafted the initial manuscript. All authors contributed to the manuscript revision and refinement. All authors reviewed and approved the final version.

\section*{Acknowledgements}
The authors are grateful for insightful discussions with J.~Actor, V.~Bouchard, M.~G.~Constantinescu, E.~Cyr, F.~Gay-Balmaz, S.~Gleyzer, T.~Griffith, J.~Hocher, D.~D.~Holm, L.~Kersten, M.~Leok, K.~Matchev, T.~Poston, M.~Serpe, A.~Sinclair, I.~K.~Tezaur, B.~Thielman, S.~Tobet, D.~Volchenkov, and D.~V.~Zenkov. We wish to express our particular gratitude to Prof. Terry Gannon; this project has greatly benefited from his continued support, active interest, and the inspiring discussions and ideas shared throughout its development.

The authors acknowledge the High Performance Computing (HPC) resources provided by the University of Alabama Office of Information Technology that contributed to the research results reported in this paper.

This work was supported in part by the New Frontiers in Research Fund (NFRF) Exploration grant NFRF-2024-00952 from the Government of Canada. VP also acknowledges the support of the Shelby Endowment at the University of Alabama. 

\bibliographystyle{unsrt}
\bibliography{references,references_MB}

\end{document}